\documentclass[]{style}

\usepackage{wrapfig}
\usepackage{tabularx}
\usepackage{textcomp}
\usepackage{stfloats}
\usepackage{url}
\usepackage{verbatim}
\usepackage{titlesec}
\usepackage{tocloft}
\usepackage{adjustbox}
\usepackage{multirow}
\usepackage{pifont}
\usepackage{tikz}
\usepackage{comment}
\usepackage{amsmath,amssymb} 
\usepackage{colortbl}  
\usepackage{xcolor}
\usepackage{booktabs} 
\usepackage{natbib}
\usepackage{nicematrix} 
\usepackage[normalem]{ulem}
\setcitestyle{square,comma,numbers,sort&compress}
\usepackage{graphicx}    
\usepackage{subcaption} 
\usepackage{multirow} 
\usepackage{booktabs} 
\usepackage{subcaption} 
\RequirePackage{xspace}
\makeatletter
\DeclareRobustCommand\onedot{\futurelet\@let@token\@onedot}
\def\@onedot{\ifx\@let@token.\else.\null\fi\xspace}

\def\ie{\emph{i.e}\onedot}

\makeatother

\definecolor{adptorange}{RGB}{248, 205, 172}
\definecolor{cmpblue}{RGB}{189, 215, 238}
\definecolor{cmpblue}{RGB}{189, 215, 238}

\definecolor{our_red}{RGB}{232,157,160}
\definecolor{our_blue}{RGB}{136,206,230}
\definecolor{our_orange}{RGB}{246,200,168}
\definecolor{our_green}{RGB}{178,211,164}

\definecolor{attn_code0}{RGB}{247,215,200}
\definecolor{attn_code1}{RGB}{238,169,139}
\definecolor{mlp_code0}{RGB}{204,201,221}
\definecolor{mlp_code1}{RGB}{102,95,153}

\definecolor{token_blue}{RGB}{84, 120, 140}

\usepackage{amsmath,amsfonts,bm}

\def\eqref#1{equation~\ref{#1}}

\def\1{\bm{1}}

\DeclareMathAlphabet{\mathsfit}{\encodingdefault}{\sfdefault}{m}{sl}
\SetMathAlphabet{\mathsfit}{bold}{\encodingdefault}{\sfdefault}{bx}{n}

\usepackage{amsmath,amsfonts,bm}

\def\eqref#1{equation~\ref{#1}}

\def\1{\bm{1}}

\DeclareMathAlphabet{\mathsfit}{\encodingdefault}{\sfdefault}{m}{sl}
\SetMathAlphabet{\mathsfit}{bold}{\encodingdefault}{\sfdefault}{bx}{n}

\usepackage{multirow}
\usepackage{diagbox}
\usepackage{makecell}
\usepackage{tabularx}
\usepackage{graphicx}

\usepackage{array}
\usepackage{rotating}

\definecolor{aliceblue}{rgb}{0.94, 0.97, 1.0}
\definecolor{citecolor}{HTML}{0071BC}
\definecolor{linkcolor}{HTML}{ED1C24}
\definecolor{darkgreen}{HTML}{539165}

\makeatletter
\newcommand{\thickhline}{%
 \noalign {\ifnum 0=`}\fi \hrule height 1pt
 \futurelet \reserved@a \@xhline
}
\makeatother

\usepackage{pifont}
\newcommand{\one}{\textcolor{violet}{\ding{182}}}
\newcommand{\two}{\textcolor{violet}{\ding{183}}}
\newcommand{\three}{\textcolor{violet}{\ding{184}}}

\newcommand{\listnumber}[1]{\textbf{\color{violet}{#1}}}

\usepackage{pifont}       
\usepackage{bbding}       
\usepackage{fontawesome}
\usepackage{xspace}

\usepackage{float}
\usepackage{enumitem}

\newlength\savewidth

\newcolumntype{x}[1]{>{\centering\arraybackslash}p{#1pt}}
\newcolumntype{y}[1]{>{\raggedright\arraybackslash}p{#1pt}}
\newcolumntype{z}[1]{>{\raggedleft\arraybackslash}p{#1pt}}

\renewcommand{\paragraph}[1]{\vspace{1mm}\noindent\textbf{#1}}
\usepackage{colortbl}
\usepackage{xcolor}
\usepackage{wrapfig}

\renewcommand{\paragraph}[1]{\vspace{1.25mm}\noindent\textbf{#1}}

\newcommand{\modelname}{Penguin\xspace}

\usepackage{algorithm}
\usepackage{listings}

\definecolor{codeblue}{rgb}{0.25, 0.5, 0.5}
\definecolor{codekw}{rgb}{0.35, 0.35, 0.75}
\lstdefinestyle{Pytorch}{
    language = Python,
    backgroundcolor = \color{white},
    basicstyle = \fontsize{9pt}{8pt}\selectfont\ttfamily\bfseries,
    columns = fullflexible,
    aboveskip=1pt,
    belowskip=1pt,
    breaklines = true,
    captionpos = b,
    commentstyle = \color{codeblue},
    keywordstyle = \color{codekw},
}

\definecolor{colSubject}{HTML}{D32F2F}   
\definecolor{colAction}{HTML}{F57C00}    
\definecolor{colDetail}{HTML}{388E3C}    
\definecolor{colSpatial}{HTML}{1976D2}   
\definecolor{colMood}{HTML}{7B1FA2}      
\definecolor{colKnow}{HTML}{AFB42B}      

\usepackage{xcolor}

\usepackage{tikz}
\usetikzlibrary{tikzmark, calc, shadows.blur, shapes.geometric, fit, positioning}
\usepackage{xparse}
\pgfdeclarelayer{background}
\pgfdeclarelayer{floatbox}
\pgfdeclarelayer{foreground}
\pgfsetlayers{background,floatbox,main,foreground}

\newcounter{hcellcount}

\NewDocumentCommand{\hctext}{m}{\csname hctext@#1\endcsname}
\NewDocumentCommand{\sethctext}{mm}{\expandafter\gdef\csname hctext@#1\endcsname{#2}}

\NewDocumentCommand{\hc}{m}{%
  \stepcounter{hcellcount}%
  \sethctext{\thehcellcount}{#1}%
  \tikz[baseline=(hc-\thehcellcount.base), remember picture]{
    \node[inner sep=0pt, outer sep=0pt, anchor=base] (hc-\thehcellcount) {#1};
  }%
}

\definecolor{scoreRed}{RGB}{200, 0, 0}
\definecolor{grayText}{RGB}{120, 120, 120}

\definecolor{green}{HTML}{009000}
\definecolor{red}{HTML}{ea4335}

\definecolor{cvblue}{rgb}{0.15, 0.45, 0.68}

\title{\includegraphics[height=30pt]{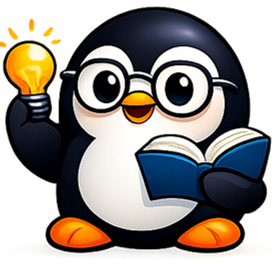}\hspace{5pt} Penguin-VL: Exploring the Efficiency Limits of VLM with LLM-based Vision Encoders}

\author[*]{Boqiang Zhang}
\author[*,\dagger]{Lei Ke}
\author[*]{Ruihan Yang}
\author[*]{Qi Gao}
\author[*, \S]{Tianyuan Qu}
\author{Rossell Chen}
\author[\ddag]{Dong Yu}
\author[\ddag]{Leoweiliang}

\affiliation[]{Penguin-VL team at Tencent AI Lab\\}

\contribution[*]{Core contributors}
\contribution[\dagger]{Proj Lead}
\contribution[\ddag]{Senior Leads}

\abstract{
Vision Language Model (VLM) development has largely relied on scaling model size, which hinders deployment on compute-constrained mobile and edge devices such as smartphones and robots. In this work, we explore the performance limits of compact (e.g., 2B and 8B) VLMs.  We challenge the prevailing practice that state-of-the-art VLM must rely on vision encoders initialized via massive contrastive pretraining (e.g., CLIP/SigLIP). We identify an objective mismatch:
contrastive learning, optimized for discrimination, enforces coarse and category-level invariances that suppress fine-grained visual cues needed for dense captioning and complex VLM reasoning. To address this issue, we present \textbf{\modelname-VL}, whose vision encoder is initialized from a text-only LLM. Our experiments reveal that \modelname-Encoder serves as a superior alternative to traditional contrastive pretraining, unlocking a higher degree of visual fidelity and data efficiency for multimodal understanding. Across various image and video benchmarks, \modelname-VL achieves performance comparable to leading VLMs (e.g., Qwen3-VL) in mathematical reasoning and surpasses them in tasks such as document understanding, visual knowledge, and multi-perspective video understanding. Notably, these gains are achieved with a lightweight architecture, demonstrating that improved visual representation\textemdash rather than model scaling\textemdash is the primary driver of performance. Our ablations show that \modelname-Encoder consistently outperforms contrastive-pretrained encoders, preserving fine-grained spatial and temporal cues that are critical for dense perception and complex reasoning. This makes it a strong drop-in alternative for compute-efficient VLMs and enables high performance in resource-constrained settings.

Code: \url{https://github.com/tencent-ailab/Penguin-VL} \\
2B Model: \url{https://huggingface.co/tencent/Penguin-VL-2B} \\
8B Model: \url{https://huggingface.co/tencent/Penguin-VL-8B} 
}

\date{\today} 
\makeatletter
\def\blfootnote{\gdef\@thefnmark{}\@footnotetext}
\makeatletter

\begin{document}

\thispagestyle{firstheader}
\blfootnote{$^\S$Work done during an internship at Tencent AI Lab.}

\maketitle
\pagestyle{empty}

\section{Introduction}

\begin{figure}[!htbp]
     \centering
     \begin{subfigure}[b]{0.48\textwidth}
         \centering
         \includegraphics[width=\textwidth,
  trim=0cm 1cm 0cm 0cm,
  clip]{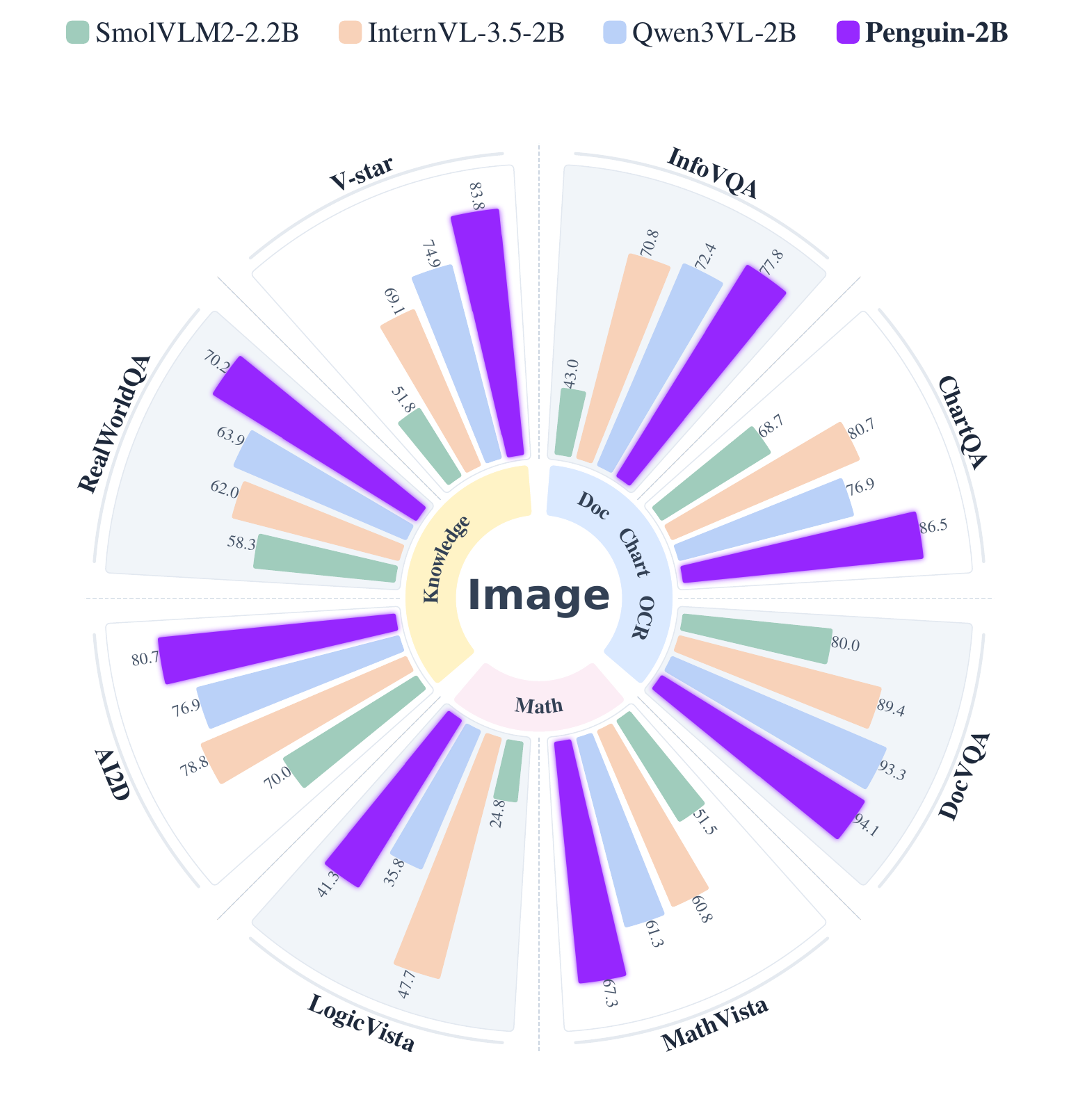}
         \label{fig:radar_a}
     \end{subfigure}
     \hfill
     \begin{subfigure}[b]{0.48\textwidth}
         \centering
         \includegraphics[width=\textwidth,
  trim=0cm 1cm 0cm 0cm,
  clip]{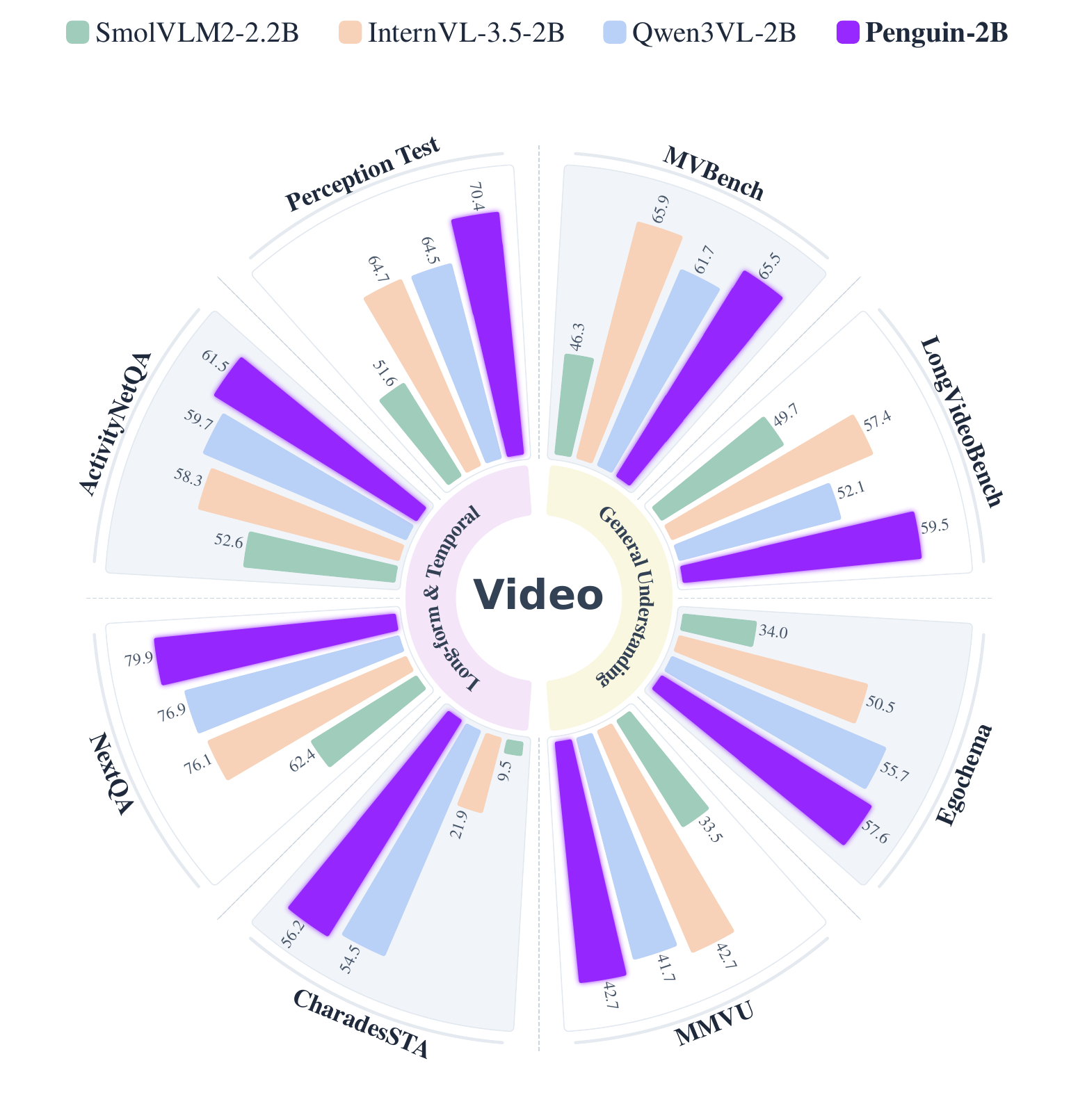}
         \label{fig:radar_b}
     \end{subfigure}
     \caption{\textbf{Benchmark comparison across tasks.}
\textbf{Left (Image):} Image benchmarks grouped by capability: \textbf{OCR}, \textbf{Math}, and \textbf{Knowledge}.
\textbf{Right (Video):} Video benchmarks grouped into \textbf{Long-form \& Temporal} and \textbf{General Understanding}.
Each sector corresponds to a benchmark, and the radial bars indicate relative performance, where longer bars mean better results.
Overall, \textbf{Penguin-VL 2B} achieves excellent performance across all modalities, highlighting clear advantages over existing state-of-the-art opensource models.}
     \label{fig:overall_radar}
\end{figure}

The integration of robust visual perception into Large Language Models (LLMs) has yielded extraordinary multi-modal capabilities. Recent advances, such as Qwen3-VL~\cite{bai2023qwen, chu2024qwen2, qwen3technicalreport}, Intern-VL~\cite{chen2023internvl,internvl2.5,wang2025internvl3}, and Molmo~\cite{deitke2025molmo,deitke2025molmo} series, demonstrate that coupling strong LLMs with modality-specific encoders~\cite{tschannen2025siglip,radford2021learning} can yield impressive general-purpose capabilities. However, a significant gap remains between state-of-the-art research prototypes and practical deployment. In real-world scenarios, the requirement is rarely just ``the strongest model possible'' at any cost; rather, it is for a model that is \emph{compact enough to serve efficiently} while remaining \emph{reliably strong} across modalities. 
Currently, leading Vision Language Models (VLMs) often rely on massive parameter counts and heavy training pipelines. This dependence complicates deployment under strict latency constraints and frequently results in uneven performance, such as when a model optimized for image understanding struggles with video temporal reasoning.

This technical report targets a pragmatic goal: \textbf{building a compact vision-centric multi-modal foundation model with consistently strong capability across images and videos}. We posit that the bottleneck for efficient multimodal systems lies not just in the LLM or training data, but in the visual backbone itself. Therefore, we do not simply propose another model; rather, we introduce a holistic framework centered on visual representation learning. Our approach involves: (1) \textit{a novel vision encoder and pretraining strategy} specifically engineered for VLM systems by leveraging modern LLM architectural principles; (2) \textit{a tailored vision language model and training recipe} that progressively harmonizes fine-grained perception with cross-modal reasoning; and (3) \textit{a comprehensive data curation pipeline} designed to prepare our training data. To support our claims, we show that our method yields best-in-class performance among parameter-efficient VLMs (2B and 8B size). Furthermore, our encoder comparison studies explicitly validate that our custom vision encoder and curricula are the primary drivers of these efficiency gains (Section~\ref{sec:exp}).

\paragraph{Revisiting the Vision Encoder: The Motivations}
In recent years, the prevailing paradigm for training modern VLMs~\citep{Bai2025Qwen3VLTR,internvl2.5,wang2025internvl3,bai2025qwen2,guo2025seed1} has been to initialize the vision encoder using large-scale contrastive pretraining~\citep{radford2021learning,Zhai2023SigmoidLF}. However, when CLIP~\citep{radford2021learning} was first introduced in 2021, there was limited systematic investigation or empirical evidence establishing contrastive learning as an optimal pretraining strategy for vision encoders in VLMs. In practice, transformers trained with contrastive objectives typically adopt a discriminative training paradigm, where supervision is applied only to a global summarization token (e.g., a \texttt{[CLS]} token or attention pooling). This objective is fundamentally misaligned with the sequence token prediction formulation that underpins language modeling. While recent efforts like SigLIP 2~\citep{tschannen2025siglip} attempt to mitigate this via caption-like objectives, they largely retain vision-centric-only inductive biases and ViT architecture~\citep{dosovitskiyimage}. Consequently, there remains significant potential in exploring alternative initialization strategies that can better align with language generative tasks. Specifically, two promising avenues have yet to be fully realized in vision encoding: leveraging \textit{language pretraining knowledge} to imbue the encoder with rich semantic priors and reasoning capabilities from the outset, as language models and adopting the \textit{efficient transformer architecture design} of modern LLMs, which is natively optimized for scalable, dense sequence modeling. This perspective is reinforced by parallel advancements in speech modeling, where recent approaches successfully fine-tune text-only LLMs to natively process and generate continuous speech signals~\citep{peng2025vibevoice,hu2026qwen3}.

\paragraph{A Comprehensive Training Recipe}
Adapting a text-based LLM to serve as a vision encoder, however, presents distinct challenges. Visual patches require non-causal processing, and bridging language-only embeddings to capture visual-language correlations requires targeted architectural adjustments. To ensure stable optimization, our framework introduces two key modifications: (1) adapting the attention mechanism and positional encoding to process vision tokens bidirectionally; (2) implementing a novel distillation loss during a warm-up phase to smoothly inject visual knowledge to a text model. Section~\ref{sec:ablation} details our ablations validating the effectiveness of these choices.

Building upon this robust encoder, our complete VLM training recipe begins with joint pretraining on a high-quality, diverse mixture of natural images, documents, and charts. We then apply a two-stage Supervised Fine-Tuning (SFT) strategy: initially establishing general compliance using a blend of image and video instructions, before pivoting to complex video-centric reasoning tasks. Finally, to efficiently process extended video sequences, we introduce a Temporal Redundancy-Aware (TRA) token compression mechanism. By distinguishing keyframes from intermediate contexts, TRA dynamically optimizes the token budget—progressively reducing redundancy while preserving critical temporal dynamics.

\begin{figure}[!t]
\centering
\vspace{-0.2in}
\includegraphics[width=\linewidth]{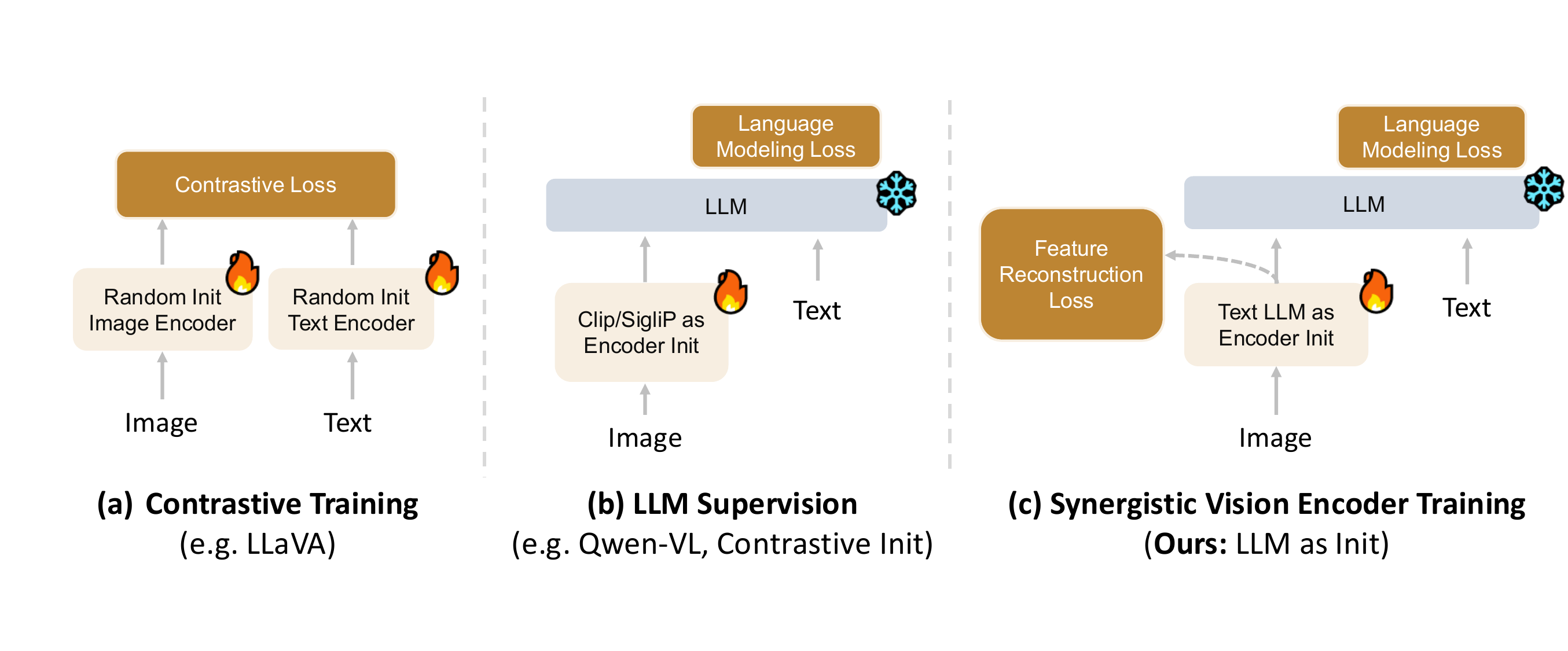}

\caption{\textbf{Comparison of Vision Encoder Training Paradigms for VLMs.}. We contrast three distinct approaches for training the vision encoder: \textbf{(a) Contrastive Training}, which relies on contrastive loss between image and text encoders and then frozen the vision encoder during VLM training. While the training logic is simple, this paradigm requires massive datasets, suffers from training instability, and often yields insufficient fine-grained multimodal alignment. \textbf{(b) Direct LLM Supervision} initializes the vision encoder with contrastive pre-trained weights and aligns visual features to a frozen LLM (via language modeling loss). 
While this enables direct alignment with the LLM feature space, it is highly sensitive to data quality and prone to overfitting to the training image distribution. \textbf{(c) \modelname-Encoder Training (Ours)}, which fuses the advantages of previous methods by initializing the vision encoder directly with the weights of a Text LLM. This approach ensures the starting model distributions are close—facilitating easier alignment—while equipping the vision model with rich initial linguistic knowledge and enabling simple, efficient scaling of vision parameters. 
}
\label{fig:method_compare}
\end{figure}

We summarize the key contributions of this work as follows:
\begin{itemize}
    \item \textbf{Encoder.} We propose \modelname-encoder, a new vision encoder directly adapted from a text-only LLM architecture. As illustrated in Figure~\ref{fig:method_compare}, our design fundamentally departs from mainstream vision encoders by reusing the LLM backbone weights, enabling tighter modality alignment and advanced architectural improvement.

    \item \textbf{Mixed Supervision Encoder Pretraining.} We introduce a dedicated auxiliary objective tailored to the proposed encoder, allowing effective joint utilization of large-scale labeled and unlabeled structured data (e.g., charts). This mixed supervision strategy substantially improves data efficiency and representation quality during early-stage pretraining.

    \item \textbf{Unified Training Recipe.} We present a comprehensive \modelname-VL training pipeline that integrates a low-to-high resolution curriculum, priority-aware video token compression, and a two-stage instruction-tuning strategy that harmonizes image and video capabilities.

    \item \textbf{Strong Performance at Compact Scale.} Our approach achieves consistently strong results across standard image and video benchmarks while maintaining a compact and computationally efficient model sizes.
\end{itemize}

\section{Methodology}

\begin{figure}[t]
\centering
\includegraphics[width=\linewidth,
  trim=0cm 0cm 0cm 0cm,
  clip]{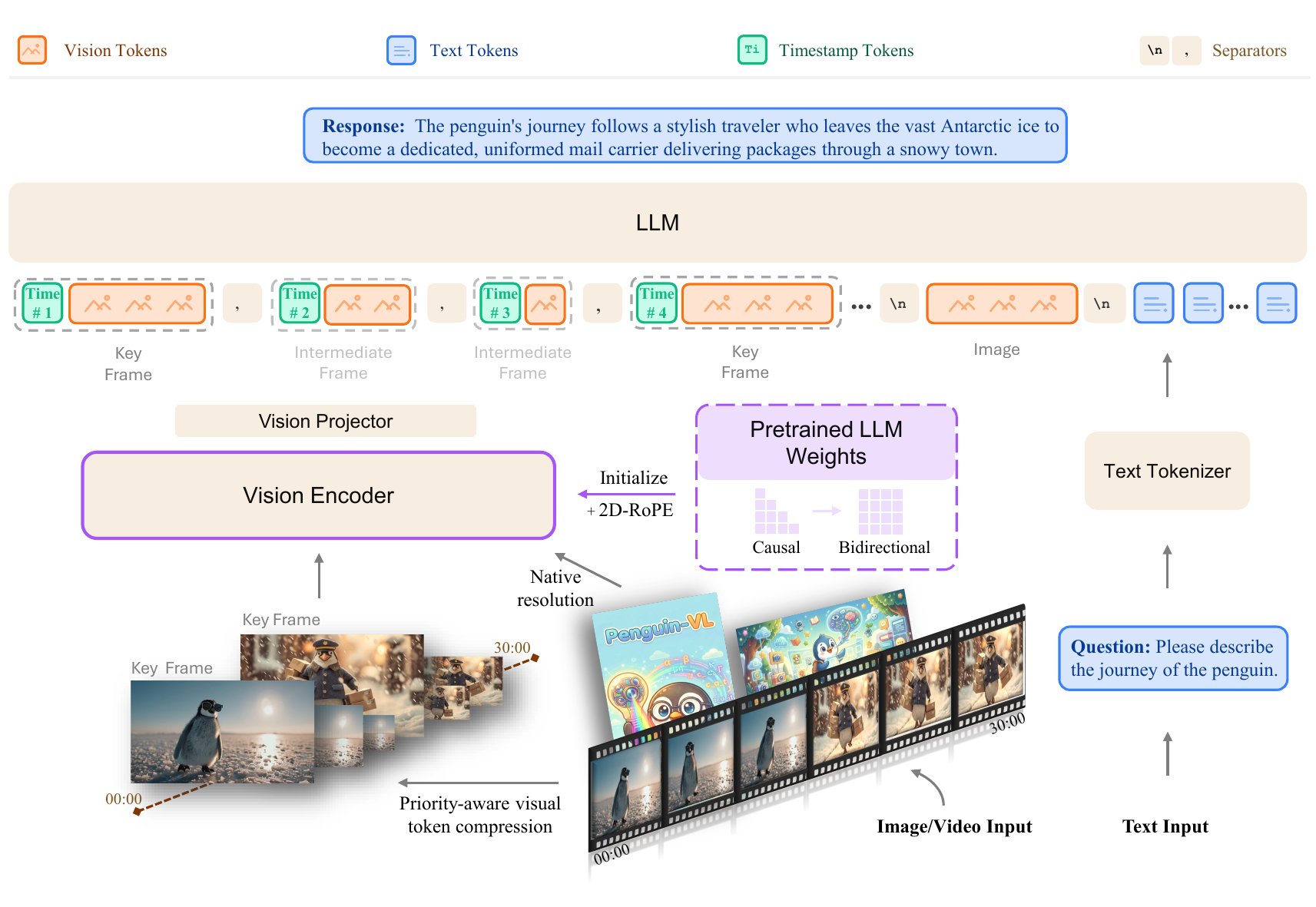}
\caption{\textbf{Method Overview.} 
\modelname-VL adopts a unified architecture for vision understanding. 
\textbf{Vision:} The vision encoder is initialized from a text-only LLM (Qwen3-0.6B) and equiped with 2D-RoPE and bidirectional attention. To handle long video contexts efficiently, we employ a \textit{Temporal Redundancy-Aware (TRA)} strategy that dynamically allocates token budgets among key frames and intermediate frames. }

\label{fig:method_overview}
\end{figure}

\subsection{Model Architecture}

Our \modelname series comprises \modelname-Encoder and \modelname-VL, built upon Qwen3 LLM backbones~\cite{qwen3technicalreport} and available in 2B and 8B parameter variants (\modelname-Encoder maintains a fixed parameter size, independent of the VLM backbone.). To achieve well-rounded understanding, \modelname adopts a unified three-module design consisting of (i) a LLM-based vision encoder (\modelname-Encoder), (ii) an MLP-based vision--language merger/projector, and (iii) a LLM. Figure~\ref{fig:method_overview} illustrates the overall architecture.

\subsubsection{\modelname-Encoder: From a Text-only LLM to Vision Encoder}\label{sec:vision_encoder}

Modern LLMs employ more advanced architectures for information processing and are pretrained at scale on massive corpora containing rich semantic and factual knowledge. This observation motivates an alternative perspective: rather than learning visual representations entirely from scratch, one can leverage the world knowledge already embedded in an LLM and focus on learning the alignment between visual concepts and the textual knowledge it already possesses.

Following this insight, unlike Qwen3-VL~\cite{Bai2025Qwen3VLTR}, which adopts a SigLIP-2–style vision backbone \citep{tschannen2025siglip} and continues training from publicly released vision checkpoints, we initialize our vision encoder directly from a text-only large language model, specifically Qwen3-0.6B~\cite{qwen3technicalreport}. Concretely, \textit{we transform the LLM’s causal self-attention into bidirectional full attention}, enabling the symmetric token interactions required for effective visual representation learning. To support variable-resolution inputs, we further equip the encoder with 2D rotary positional embeddings (2D-RoPE), as commonly adopted in recent multimodal models such as VideoLLaMA~\cite{DBLP:conf/emnlp/ZhangLB23,cheng2024videollama,zhang2025videollama} series and Qwen-VL series~\cite{wang2024qwen2,bai2025qwen2,Bai2025Qwen3VLTR}. Inputs are processed at their original spatial resolution whenever the token budget permits; otherwise, they are resized appropriately. This design ensures stable behavior across a wide range of aspect ratios and spatial scales.

This text-to-vision initialization strategy offers four distinct practical advantages: (i) \textbf{Architectural Expressivity}: Unlike standard SigLIP encoders, LLM-based architectures incorporate advanced design elements such as QK normalization~\cite{henry2020query}, which significantly enhance feature stability and expressiveness; (ii) \textbf{Native Alignment}: The vision encoder begins in a representation space inherently compatible with the downstream LLM decoder, thereby minimizing the modality gap; (iii) \textbf{Semantic Priors}: It inherits rich linguistic knowledge from the outset, providing strong semantic grounding for visual concepts; and (iv) \textbf{Scalability}: It facilitates predictable scaling by leveraging established LLM design principles and parameterization. If required by future scenarios, this framework can be readily extended to train vision encoders with substantially larger model capacities (e.g., 1.5B parameters), compared to alternative approaches such as contrastive learning. More ablation studies and discussions can be found in Section \ref{sec:ablation}.

To better align the LLM-initialized vision encoder with visual concepts and the visual feature space, we propose a training paradigm that jointly employs LLM cross-entropy supervision and reconstruction-based objectives. In the first stage of vision encoder training, we introduce a reconstruction (distillation) loss to encourage the model to preserve fine-grained visual details as well as the overall distribution of visual features. The reconstruction loss contain three parts:

(1) Amplitude Loss: Assume that the features produced by the vision encoder are denoted as $F_{s}$, and the corresponding teacher supervision features are denoted as $F_{t}$. The amplitude loss directly supervises the absolute value of $F_{s}$:

$$
L_{A}=\frac{1}{N}\sum_N(|F_{s}-F_{t}|)
$$

(2) Direction Loss: Since the amplitude loss is prone to overfitting and representation collapse, we introduce a directional loss to better align the LLM-initialized feature distribution with the visual feature space. The direction loss is set as a consine similarity:

$$
L_{D}=\frac{1}{N}\sum_N\left(tr\left(\frac{F_sF_t^\top}{||F_s||_2||F_t||_2}\right)\right)
$$

(3) Relation Loss: Based on the two losses above, effective supervision can already be achieved. However, motivated by the unique properties of the attention mechanism, we further propose a third objective, namely a relational loss. In attention-based models, the focus is placed on the relationships between tokens rather than the absolute attributes of individual tokens (such as magnitude or direction). Moreover, for vision models, the interactions among different patches are crucial for modeling the underlying visual space. To this end, we formulate the relational loss by leveraging self-correlation similarity to explicitly supervise inter-patch relationships, and we validate its effectiveness in Table~\ref{tab:ablation}:

$$
L_{R}=\frac{1}{N}\sum_N\left(\left|\frac{F_sF_s^\top}{||F_s||_2^2}-\frac{F_tF_t^\top}{||F_t||_2^2}\right|\right)
$$

\subsubsection{Video Encoding and Projector}
\label{sec:pac}

\paragraph{Temporal Redundancy-Aware Visual Token Encoding and Compression.}
With our fine-tuned vision encoder, we first partition the visual input (image or video) into patches and map them to token embeddings through a linear projection layer. The resulting visual tokens are then processed by the vision encoder to produce visual features. For image inputs, when the token length exceeds the maximum allowable context, we down-scale the image to fit the budget. 
For video inputs, we employ a flexible temporal sampling strategy: we sample at a fixed sample rate, which is then capped at a maximum number of frames ($max\_frames$) per video. If the video exceeds $max\_frames$, we switch to uniform sampling, extracting a fixed set of $max\_frames$ spaced equidistantly across the entire video.

In contrast to Qwen3-VL, which adjusts the global input resolution based on the token limit while maintaining a consistent resolution across all frames, our design explicitly prioritizes \emph{content-adaptive granularity}. Specifically, we introduce \emph{Temporal Redundancy-Aware token compression} (TRA) strategy to allocate more tokens to the most informative frames while using fewer for less informative ones. This allocation is governed by a four-stage comprehensive policy that balances information preservation, temporal consistency, and the physical lower bound on spatial resolution. TRA classifies frames based on temporal similarity into \textit{key frames} (capturing rapid temporal changes) and \textit{intermediate frames} (providing stable context). A critical advantage of this design is its robustness to resolution variation: since our vision encoder and VLM are jointly pretrained over a wide range of image resolutions, the model supports continuous resolution adaptation. This facilitates smooth, dynamic down-scaling, effectively preserving inter-frame spatial continuity even under aggressive compression.

Let $T_{\max}$ denote the global visual-token budget, and $T_{\min}$ be the minimum allowable token count per frame to preserve semantic integrity. We denote the per-frame token counts for key frames and intermediate frames as $T_k$ ($k \in \mathcal{K}$) and $T_i$ ($i \in \mathcal{I}$), where $\mathcal{K}$ and $\mathcal{I}$ denote the sets of indices for key and intermediate frames, respectively. TRA proceeds in a three-stage cascade:

\begin{itemize}
    \item \textit{Compression Stage 1} (Resolution Preservation): 
    In the initial stage, our primary objective is to maximize the information retention of the original video. If the full-resolution tokenization of the input video satisfies
    \begin{equation}
    \sum_{k \in \mathcal{K}} T_k \;+\; \sum_{i \in \mathcal{I}} T_i \;\le\; T_{\max},
    \label{eq:budget}
    \end{equation}
    all frames are tokenized at their native resolutions without applying any spatial compression. This strategy is particularly critical for short videos characterized by rapid temporal pacing. In such cases, maintaining high spatial resolution is effective in capturing subtle action dynamics and fine-grained scene details.

    \item \textit{Compression Stage 2} (Synchronous Downscaling): 
    If the total token count under full-resolution tokenization exceeds $T_{\max}$, we proceed to the synchronous downscaling process. In this process, both key and intermediate frames are down-scaled while maintaining a fixed relative ratio (specifically, intermediate frames maintain a $4\times$ spatial down-sampling relative to key frames, implying $T_k \approx 16 T_i$). We utilize bilinear interpolation to resize the frames by applying a continuous scaling factor $\alpha \in (0,1]$, updating the token counts as follows until the budget constraint in Eq.~(\ref{eq:budget}) is met:
    \begin{equation}
    T_k \leftarrow \alpha T_k, \qquad
    T_i \leftarrow \alpha T_i.
    \end{equation}
    Joint continuous scaling preserves cross-temporal spatial consistency and ensures that the visual quality of fast-motion (key) and slow-context (intermediate) frames degrades evenly across the video sequence.

    \item \textit{Compression Stage 3} (Saturation-Aware Scaling): 
    As compression intensifies, intermediate frames may reach the physical lower bound $T_i = T_{\min}$. Upon reaching this threshold, we clamp intermediate frames at $T_{\min}$ and transfer all remaining compression pressure exclusively to the key frames. We continue to down-scale $T_k$ until the global budget is satisfied. This mechanism prevents intermediate frames from being over-compressed, ensuring the preservation of essential semantic context and structural details. Finally, the compression process concludes once the key frames are reduced sufficiently to satisfy Eq.~(\ref{eq:budget}). Our system design guarantees that, given our setting of $T_{\max}$, the token count for key frames $T_k$ will never breach the floor $T_{\min}$ even at the maximum supported duration ($max\_frames$).
\end{itemize}

\paragraph{MLP-based Vision--Language Projector.}
To align visual features with the language model, we employ a lightweight MLP-based vision--language projector.
Unlike prior designs~\cite{yu2025minicpm} that perform spatial token compression, our merger directly projects the visual feature dimension
to match the LLM hidden size through a simple feed-forward transformation.
This design choice prioritizes simplicity and efficiency, enabling seamless integration of visual tokens into the LLM
without introducing additional spatial restructuring or complex merging operations.

\subsection{Construction of High-Quality Multi-modal Recaption and QA Datasets}

We construct a large-scale, high-quality multimodal corpus of images and videos for model training. The data construction pipeline follows a \textbf{quality- and diversity-first} principle and consists of three stages: (1) large-scale source aggregation, (2) multi-stage filtering and de-duplication, and (3) scalable automatic annotation using strong captioning and annotation models. This corpus covers a wide range of domains, uses consistent data formatting, and provides reliable supervision signals. It serves as a core component of the overall data mixture.

We use separate pipelines for image and video data. The image and video datasets are named \modelname-Recap-I and \modelname-Recap-V, respectively. The video QA dataset is named \modelname-QA. For videos, we additionally derive question–answer (QA) pairs from recaptioning results. We next describe the construction processes for image and video data, respectively.

\subsubsection{Image Data Curation}\label{sec:image_data_curation}

\paragraph{Quality filtering and diversity balancing.}
To construct our \modelname-Recap-I image dataset, which consists of 57.2 million image and text pairs, we first sample large-scale image data from COYO-700M~\citep{kakaobrain2022coyo-700m} and DataComp-1B~\citep{gadre2023datacomp}. We then adopt a two-stage data curation pipeline that emphasizes both quality control and diversity preservation. In the first stage, we filter out low-quality and irrelevant images using basic visual attributes, including resolution, aspect ratio, and corruption indicators, thereby removing samples that are severely blurred, extremely low-resolution, or corrupted. In the second stage, we promote semantic diversity through a scalable hierarchical clustering strategy. Specifically, we extract CLIP embeddings for all images and perform semantic clustering using the k-means algorithm. To mitigate the prohibitive computational cost of clustering at full scale, we introduce a hierarchical variant of k-means. We first apply k-means to a randomly sampled subset to obtain coarse cluster centroids, assign all images to their nearest centroids, and then recursively perform finer-grained clustering within each sub-cluster. After clustering, we employ a greedy selection strategy within each cluster to choose the top-$N$ samples that are maximally separated in the embedding space. This procedure encourages intra-cluster diversity while preventing the over-representation of dominant concepts, resulting in a more balanced and diverse image distribution across semantic categories and visual styles.


\paragraph{Detailed Long Captioning for richer supervision.}
For each unlabeled or weakly labeled image, we first perform a comprehensive structured analysis by explicitly annotating multiple fine-grained aspects of the visual content. Concretely, the model is prompted to identify and describe:

\begin{tcolorbox}[
  breakable,
  colback=gray!5,        
  colframe=gray!40,      
  boxrule=0.5pt,         
  arc=2pt,               
  left=8pt, right=8pt,   
  top=8pt, bottom=8pt
]
    \begin{description}[style=unboxed, leftmargin=0pt, itemsep=0.8em, font=\bfseries\color{black!80}]
        
        \item[Global Image Semantics.] 
        A concise yet descriptive title and an overall summary capturing the primary scene, salient events, and high-level semantics depicted in the image.

        \item[Subjects.] 
        All visible subjects (e.g., people, animals, etc.), along with detailed visual descriptions encompassing their appearance, pose, clothing, material, color, and relative prominence.

        \item[Actions and Activities.] 
        Ongoing actions, interactions, or events involving the identified subjects, including both individual activities and multi-entity interactions.

        \item[Spatial Relationships.] 
        Explicit spatial configurations and relationships among entities, such as relative positions, distances, orientations, containment, and hierarchical layouts within the scene.

        \item[Scene and Contextual Attributes.] 
        Environmental and contextual cues, including location or setting type, time of day, weather conditions (if visible), and other scene-level attributes that provide situational context.

        \item[Notable Objects and Functional Details.] 
        Salient objects or items of interest, together with their detailed descriptions, functional roles, usage contexts, and any distinctive or uncommon characteristics.

        \item[Visible Text and OCR Content.] 
        All readable text appearing in the image, including signs, labels, captions, or interface elements, along with their semantic roles and functions within the scene.

        \item[Image Quality and Perceptual Properties.] 
        Observations regarding image quality, such as clarity, resolution, lighting conditions, focus, occlusions, and potential visual artifacts.

        \item[Overall Mood or Tone.] 
        The perceived atmosphere or affective tone of the image (e.g., calm, tense, joyful), inferred from visual cues such as lighting, color, composition, and subject expressions.

        \item[Knowledge-Intensive and Reasoning-Oriented Descriptions.] 
        High-level interpretations that require external or commonsense knowledge, including inferred intent, likely activities, implied events, or contextual explanations beyond directly observable pixels.

    \end{description}
\end{tcolorbox}

After completing the structured annotations, we employ a proprietary model to synthesize a single comprehensive long-form caption by integrating all available cues, including visual attributes, spatial relationships, OCR-visible text, and relevant world knowledge. The resulting caption is designed to be maximally informative, precise, and faithful, capturing fine-grained details of entities, actions, interactions, and environmental context while avoiding redundancy. This aggregation step converts dense structured supervision into a coherent natural-language description that is well suited for large-scale generative training.

\subsubsection{Video Data Curation}
\label{sec:video_data_curation}

\paragraph{Quality Filtering and Diversity Balancing.}
To construct \modelname-Recap-V dataset, we curate 3.7M video-text pairs from 29 public video datasets. To ensure high data quality, we implement a rigorous filtering pipeline designed to eliminate both redundant duplicates and static videos. Specifically, we extract video-level features from uniformly sampled frames (e.g., 16 frames/video) and apply clustering-based deduplication to remove near-duplicates. Furthermore, to preserve clips with meaningful temporal dynamics, we compute motion scores using optical flow method) and discard segments with negligible motion. Beyond quality control, we employ adaptive resampling strategies to balance domain distribution. For massive short-video collections like VIDAL-10M~\cite{zhulanguagebind} (mostly $<20$s), we perform random sampling to create a scalable subset; conversely, for Koala-36M~\cite{wang2025koala}, we implement duration-aware sampling to ensure uniform coverage across varying video lengths, thereby enriching the timestamp distribution and capturing rare scenarios.

\paragraph{Multi-granularity video annotation.}
To equip our model with robust multi-granularity comprehension capabilities, we perform comprehensive annotation and re-labeling on various public datasets (e.g., VIDAL-10M~\cite{zhulanguagebind}, Ego4D~\cite{grauman2022ego4d}). We utilize a proprietary video annotation model with carefully designed prompts to construct a hierarchical annotation structure. This process generates annotations at three progressive levels, where each subsequent level is synthesized from the preceding information:

\begin{itemize}
    \item \textbf{Event-level atomic descriptions:} Serving as the factual foundation, these are fine-grained, timestamped captions. They focus on specific actions and objective visual facts to ensure granular grounding.
    \item \textbf{Chapter-level narrative syntheses:} Based on the sequence of event-level captions, these annotations logically group atomic events into distinct narrative chapters. They articulate the narrative function, core events, and key turning points within specific time ranges to capture the storyline's progression.
    \item \textbf{Holistic video summaries:} As the final layer of analysis, these summaries synthesize the objective facts from event captions and the structural context from chapter narratives. They provide a comprehensive interpretation, detailing the central theme, character progression, etc.
\end{itemize}

\begin{figure}[t]
\centering
\includegraphics[width=\linewidth]{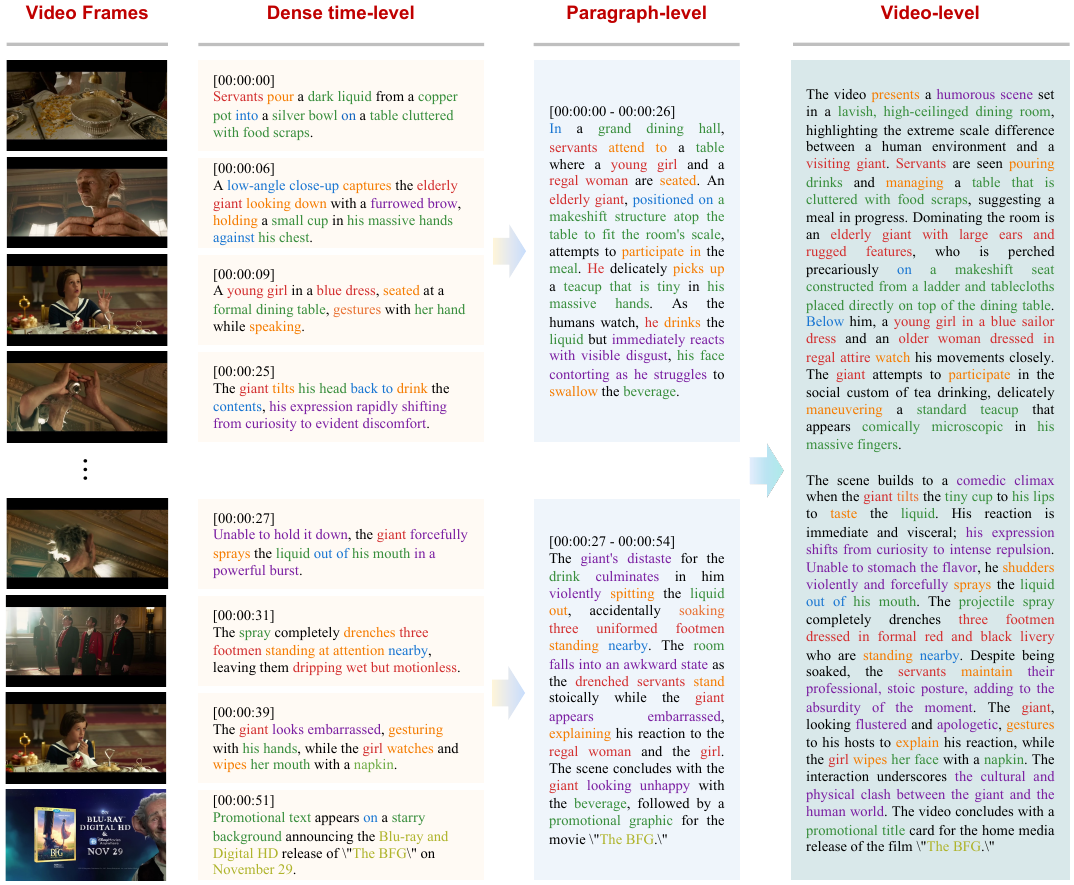}
\caption{\textbf{Multi-granularity video annotation. } This figure illustrates the alignment between visual content and textual descriptions across three temporal scales: Dense time-level, Paragraph-level, and Video-level. Semantic entities are color-coded to distinguish \textcolor{colSubject}{\textbf{Subjects}}, 
\textcolor{colAction}{\textbf{Actions}}, 
\textcolor{colDetail}{\textbf{Objects \& Details}}, 
\textcolor{colSpatial}{\textbf{Spatial Context}}, 
\textcolor{colMood}{\textbf{Mood \& Tone}}, and 
\textcolor{colKnow}{\textbf{Knowledge \& Text}}.}
\label{fig:video_dense_caption}
\end{figure}

Figure~\ref{fig:video_dense_caption} presents a visualization of these hierarchical annotations aligned with visual content. Furthermore, we run automated sanity checks to identify and discard low-quality data, specifically targeting incomplete sentences due to length truncation, chronological violations (e.g., disordered timestamps), and informational redundancy.

\paragraph{Temporal Reasoning QA Construction.} Given that the model has already established strong spatial understanding capabilities through the preceding image training stages, we aim to further enhance its temporal reasoning capabilities during instruction tuning. To this end, alongside aggregating open-source QA datasets, we curate a specialized set of temporal instruction data derived directly from our generated Dense time-level descriptions. We focus on two critical tasks: temporal ordering, where we sample multiple distinct events and prompt the model to deduce their correct chronological sequence, and temporal grounding, which requires identifying the precise start and end timestamps of specific actions. We explicitly exclude ambiguous cases (e.g., potentially overlapping events) and trivial samples, ensuring that the model learns from clear, high-quality temporal data.

\section{Training}

\subsection{Data Format}
\label{sec:data_format}

\begin{figure}[t]
\centering
\includegraphics[width=\linewidth]{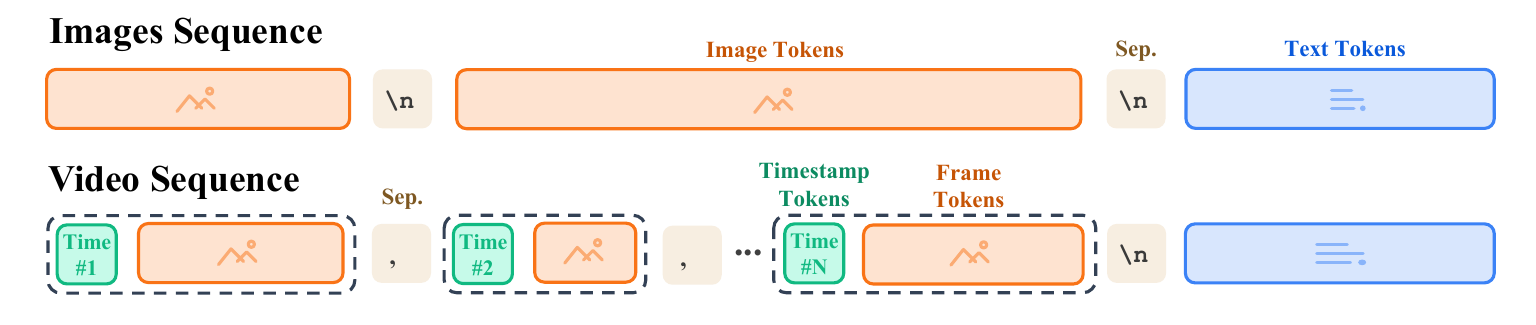}
\caption{\textbf{Data formats for different data types.} \listnumber{\one} For image sequence, we use "\textbackslash n" to separate image tokens from different image; \listnumber{\two} For video sequence, we use "Time: xxs" to indicate timestamps of each frame, "," to separate different frames, and "\textbackslash n" to separate tokens from different videos; \listnumber{\three} For streaming video sequence, videos and texts are organized in an interleaved format.}
\label{fig:data_format}
\end{figure}

We convert images and videos inputs into a single token sequence for the LLM. As detailed in Figure~\ref{fig:data_format}, the sequence is constructed using modality-specific content blocks, explicit separators, and absolute timestamp tags.

\paragraph{Notation.}
We denote the token block for an image as $\mathbf{I}$ and a video frame as $\mathbf{V}$. Text tokens are denoted by $\mathbf{X}$. To strictly align with the visual format, we use the following special tokens:
\begin{itemize}
    \item \textbf{Separators:} We use ``\texttt{\textbackslash n}'' to separate distinct inputs (e.g., images or final text) and ``\texttt{,}'' to delimit items within a continuous stream (e.g., frames).
    \item \textbf{Timestamps:} We use $\langle t \rangle$ to represent the absolute timestamp string (format: ``\texttt{Time: xxs}'').
\end{itemize}

\paragraph{Image Sequence.}
For a sequence of $N$ images, we concatenate the image blocks and separate them using ``\texttt{\textbackslash n}''. If text instructions follow, they are also separated by ``\texttt{\textbackslash n}'':
\[
\mathbf{I}_1 \;\texttt{\textbackslash n}\; \mathbf{I}_2 \;\texttt{\textbackslash n}\; \cdots \;\texttt{\textbackslash n}\; \mathbf{I}_N \;\texttt{\textbackslash n}\; \mathbf{X}.
\]

\paragraph{Video Sequence.}
A video is represented as a sequence of timestamped frames. Each frame block $\mathbf{V}_m$ is immediately preceded by its timestamp $\langle t_m \rangle$. Consecutive frames are separated by ``\texttt{,}''. The sequence ends with ``\texttt{\textbackslash n}'' before the text:
\[
(\langle t_1\rangle \mathbf{V}_1) \;\texttt{,}\; (\langle t_2\rangle \mathbf{V}_2) \;\texttt{,}\; \cdots \;\texttt{,}\; (\langle t_M\rangle \mathbf{V}_M) \;\texttt{\textbackslash n}\; \mathbf{X}.
\]

\subsection{Stage 1: \modelname-Encoder Training}
\subsubsection{Training Recipe}
We construct a dedicated vision encoder by initializing it directly from a LLM, \ie Qwen3-0.6B~\cite{qwen3technicalreport}, to inherit its parameterization and architectural priors. After removing parameters unrelated to visual processing, the model uses approximately 400M parameters for visual modeling, which is broadly consistent with SigLIP. A 2-layer MLP with GELU activation is randomly initialized as the vision projector. To ensure training stability and efficiency, we adopt a two-stage coarse-to-fine strategy. Throughout both stages, the language decoder remains frozen, and we jointly optimize the vision encoder and projector.

\paragraph{Low-Resolution Pre-training.}
In the first stage, we train on a large-scale dataset of approximately 100M samples with input resolution capped at 2048 visual tokens ($\sim600\times600$ pixels). Supervision is primarily provided by original, noisy captions.
However, relying solely on caption alignment is inefficient for structured visual domains (e.g., charts and diagrams), where natural language descriptions are often scarce or imprecise. To mitigate this and leverage large-scale unlabeled data, we augment the standard image–text loss with a reconstruction-based supervision signal via a teacher encoder. Specifically, we employ three complementary losses which are detailed in Section~\ref{sec:vision_encoder}.
Crucially, the relation loss captures the structural integrity of dense visual data, enabling us to effectively utilize unlabeled corpora to enrich visual diversity.

\paragraph{High-Resolution Fine-tuning.}
In the second stage, we remove the reconstruction branch and focus exclusively on fine-grained alignment. We increase the resolution to 10240 visual tokens and fine-tune using a filtered mixture of high-quality re-captioned data. This shift enables the model to capture detailed spatial structures and subtle semantic correspondences essential for downstream reasoning.

\subsubsection{Data Mixture}
We design the data mixture separately for the low-resolution initialization stage and the high-resolution fine-tuning stage.

During low-resolution initialization, we construct a large-scale dataset with a broad distribution to bootstrap the model’s visual perception capabilities. In addition to general image–text data, we incorporate a substantial amount of unlabeled chart and diagram data to enhance the vision encoder’s ability to handle charts and fine-grained visual patterns. To ensure wide coverage of visual concepts, we randomly sample approximately 220M image–text pairs from COYO-700M~\citep{kakaobrain2022coyo-700m} and DataComp-1B~\citep{gadre2023datacomp}. For unlabeled chart data, we curate about 2.8M samples from ChartGalaxy~\citep{li2025chartgalaxy}, M-Paper~\citep{hu2024mplug}, ChartGen~\citep{kondic2025chartgen}, and UniChart~\citep{masry2023unichart}. In total, the low-resolution initialization stage is trained on roughly 223M samples.

In the high-resolution fine-tuning stage, we increase the training resolution by allowing longer token sequences and prioritizing high-resolution visual inputs whenever feasible. At this stage, we re-annotate images using the annotation pipeline similar to Section~\ref{sec:image_data_curation}, producing detailed re-captions that provide richer semantic supervision. Specifically, we sample approximately 45M images from OpenImages~\citep{openimages}, SA-1B~\citep{kirillov2023segment}, COYO-700M~\citep{kakaobrain2022coyo-700m}, DataComp-1B~\citep{gadre2023datacomp}, and pdfa\_eng\_wds~\citep{pdfa}, and filter them based on resolution, visual clarity, and other quality criteria following procedures similar to those in Section~\ref{sec:image_data_curation}. To further broaden the data distribution and better align with video-centric scenarios, we additionally sample videos from Ego4D~\citep{grauman2022ego4d}, YouCook2~\citep{zhou2018towards}, and ShareGPT4Video~\citep{chen2024sharegpt4video}. From these videos, we randomly extract frames, treat them as image samples, and annotate them using the same pipeline, yielding roughly 2M additional samples. Overall, the high-resolution fine-tuning stage is trained on approximately 47M samples.

\subsection{Stage 2: Pre-training}
After constructing the vision encoder, we perform pre-training of the VLM. The goal of this stage is to endow the large model with diverse, broadly distributed multimodal knowledge.
During this phase, all parameters are trainable, including the LLM, the vision encoder, and the vision projector.

\subsubsection{Data Mixture}
\begin{figure*}[!t]
    \begin{minipage}[c]{0.4\textwidth}
        \includegraphics[width=\linewidth]{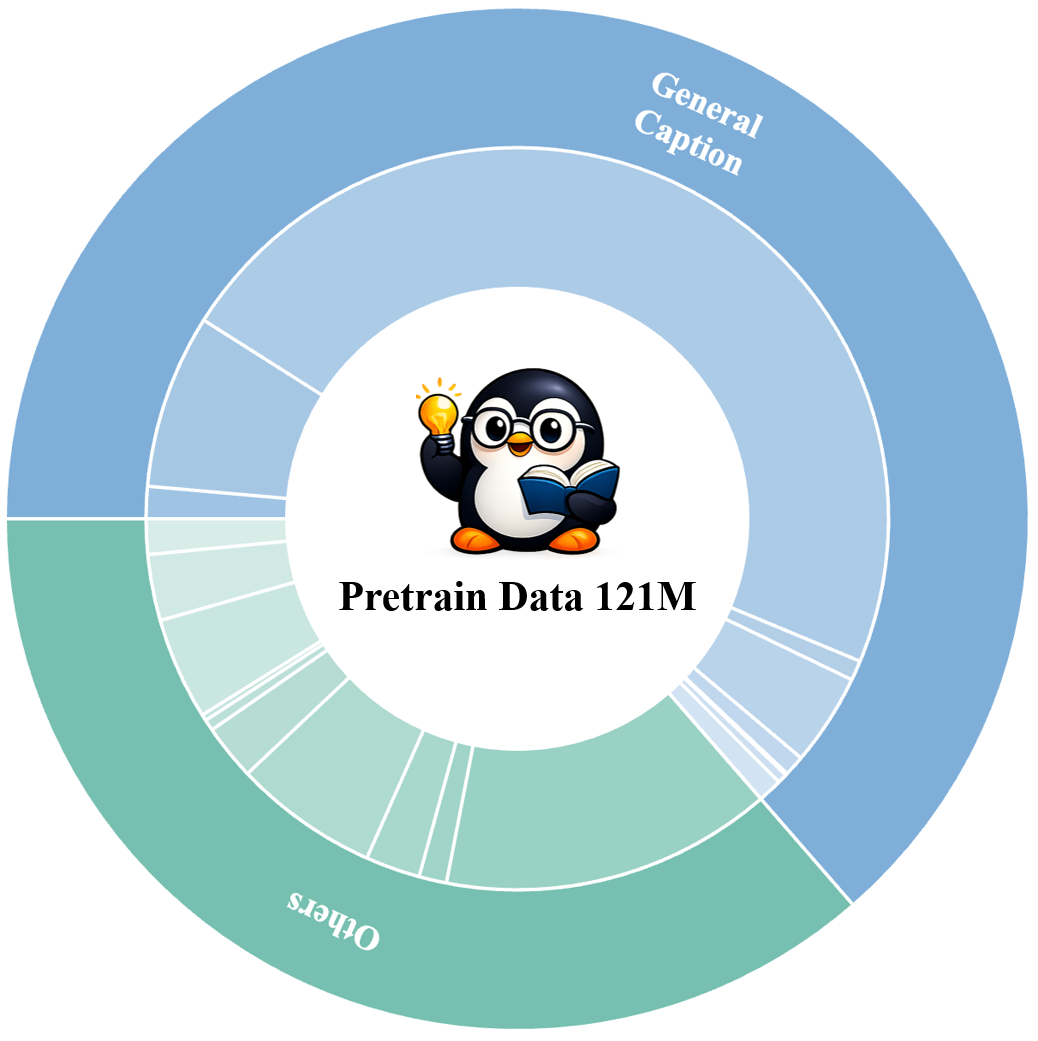}
    \end{minipage}
    \hfill
    \begin{minipage}[c]{0.55\textwidth}
        \centering
        \resizebox{0.8\textwidth}{!}{%
        \begin{tabular}{@{}ll ll@{}}
        \toprule
        \textcolor[HTML]{7FAFD9}{\rule{8pt}{8pt}} & \multicolumn{3}{l}{\textbf{{General Caption}}}\\
        \textcolor[HTML]{9fc3e3}{\rule{8pt}{8pt}} & OpenImages (1.4\%) & \textcolor[HTML]{a5c7e4}{\rule{8pt}{8pt}} & SA-1B (7.59\%)\\
        \textcolor[HTML]{accbe6}{\rule{8pt}{8pt}} & PenguinRecap-I (47.28\%) & \textcolor[HTML]{b2cfe8}{\rule{8pt}{8pt}} & DenseFusion (0.87\%)\\
        \textcolor[HTML]{b9d3ea}{\rule{8pt}{8pt}} & VL3-Syn7M (4.01\%) & \textcolor[HTML]{bfd7ec}{\rule{8pt}{8pt}} & PixmoCap (0.9\%)\\
        \textcolor[HTML]{c5dbee}{\rule{8pt}{8pt}} & UReader (0.08\%) & \textcolor[HTML]{ccdff0}{\rule{8pt}{8pt}} & FineVision (0.42\%)\\
        \textcolor[HTML]{d2e3f2}{\rule{8pt}{8pt}} & VideoFrames (1.05\%) & \textcolor[HTML]{d9e7f4}{\rule{8pt}{8pt}} & Science (0.01\%)\\
        \midrule
        \textcolor[HTML]{78C0B1}{\rule{8pt}{8pt}} & \multicolumn{3}{l}{\textbf{{Others}}}\\
        \textcolor[HTML]{9ad0c5}{\rule{8pt}{8pt}} & Document (14.45\%) & \textcolor[HTML]{a1d3c8}{\rule{8pt}{8pt}} & Region Caption (1.2\%)\\
        \textcolor[HTML]{a7d6cc}{\rule{8pt}{8pt}} & MM Code (2.38\%) & \textcolor[HTML]{aed9d0}{\rule{8pt}{8pt}} & Grounding (6.31\%)\\
        \textcolor[HTML]{b5dcd4}{\rule{8pt}{8pt}} & OCR (2.42\%) & \textcolor[HTML]{bce0d8}{\rule{8pt}{8pt}} & Interleaved (0.49\%)\\
        \textcolor[HTML]{d9e7f4}{\rule{8pt}{8pt}} & Science (0.27\%) & \textcolor[HTML]{c9e6e0}{\rule{8pt}{8pt}} & Text (4.45\%)\\
        \textcolor[HTML]{d0e9e4}{\rule{8pt}{8pt}} & Math (2.88\%) & \textcolor[HTML]{d7ece8}{\rule{8pt}{8pt}} & Code (1.54\%)\\
        \bottomrule
        \end{tabular}
        }
    \end{minipage}
    \\
    \begin{minipage}[c]{0.4\textwidth}
        \includegraphics[width=\linewidth]{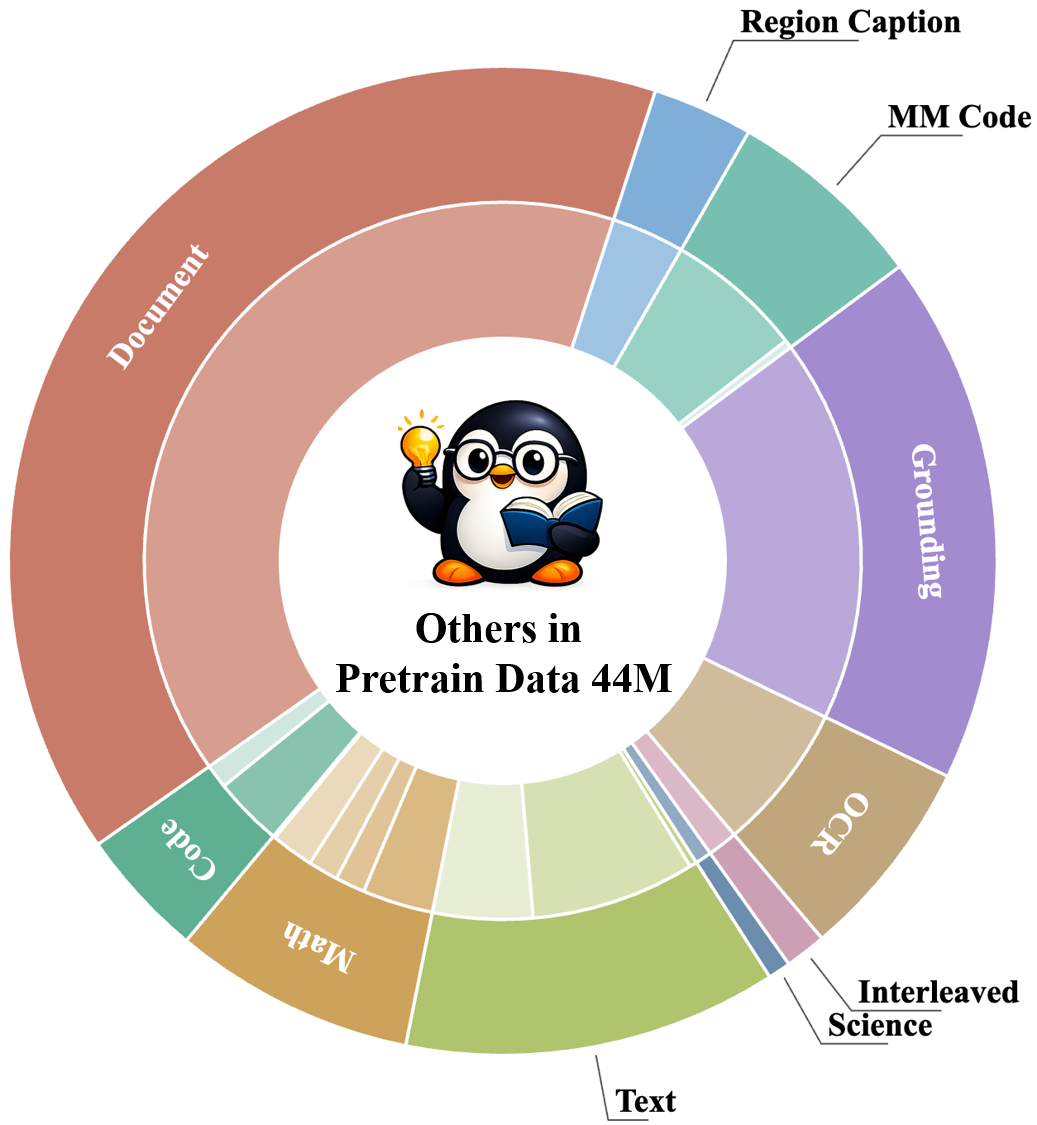}
    \end{minipage}
    \hfill
    \begin{minipage}[c]{0.55\textwidth}
        \centering
        \resizebox{\textwidth}{!}{%
        \begin{tabular}{@{}ll|ll@{}}
        \toprule
        \textcolor[HTML]{C97B6A}{\rule{8pt}{8pt}} & \textbf{Document}
        & \textcolor[HTML]{A38BCF}{\rule{8pt}{8pt}} & \textbf{Grounding} \\
        
        \textcolor[HTML]{d79c8f}{\rule{8pt}{8pt}} & PDF-A (39.72\%)
        & \textcolor[HTML]{baa8db}{\rule{8pt}{8pt}} & FGClip (17.33\%) \\
        \midrule
        
        \textcolor[HTML]{7FAFD9}{\rule{8pt}{8pt}} & \textbf{Region Caption}
        & \textcolor[HTML]{BFA67C}{\rule{8pt}{8pt}} & \textbf{OCR} \\
        
        \textcolor[HTML]{9fc3e3}{\rule{8pt}{8pt}} & Describe Anything (3.29\%)
        & \textcolor[HTML]{cfbc9d}{\rule{8pt}{8pt}} & Blip3o-OCR (6.65\%) \\
        \midrule
        
        \textcolor[HTML]{6C8DAE}{\rule{8pt}{8pt}} & \textbf{Science}
        & \textcolor[HTML]{CCA0B4}{\rule{8pt}{8pt}} & \textbf{Interleaved} \\
        
        \textcolor[HTML]{91aac2}{\rule{8pt}{8pt}} & MMSci (0.74\%)
        & \textcolor[HTML]{d9b8c7}{\rule{8pt}{8pt}} & multimodal\_textbook (1.35\%) \\
        \midrule
        
        \textcolor[HTML]{78C0B1}{\rule{8pt}{8pt}} & \textbf{MM Code}
        & \textcolor[HTML]{5FAF93}{\rule{8pt}{8pt}} & \textbf{Code} \\
        
        \textcolor[HTML]{9ad0c5}{\rule{8pt}{8pt}} & WebSight (6.19\%)
        & \textcolor[HTML]{87c3ae}{\rule{8pt}{8pt}} & Infinity\_Instruct\_code (3.13\%) \\
        
        \textcolor[HTML]{d7ece8}{\rule{8pt}{8pt}} & Chart2Code (0.37\%)
        & \textcolor[HTML]{cfe7df}{\rule{8pt}{8pt}} & KodCode (1.09\%) \\
        \midrule
        
        \textcolor[HTML]{CDA25A}{\rule{8pt}{8pt}} & \textbf{Math}
        & \textcolor[HTML]{AFC46E}{\rule{8pt}{8pt}} & \textbf{Text} \\
        
        \textcolor[HTML]{dab983}{\rule{8pt}{8pt}} & Infinity\_Instruct\_math (3.19\%)
        & \textcolor[HTML]{c3d392}{\rule{8pt}{8pt}} & Evol\_Instruct (0.32\%) \\
        
        \textcolor[HTML]{dfc496}{\rule{8pt}{8pt}} & Magpie\_pro (1.34\%)
        & \textcolor[HTML]{d5e0b3}{\rule{8pt}{8pt}} & Infinity\_Instruct\_commonsense (7.41\%) \\
        
        \textcolor[HTML]{e5cea8}{\rule{8pt}{8pt}} & Magpie\_ultra (1.36\%)
        & \textcolor[HTML]{e7edd4}{\rule{8pt}{8pt}} & smollm\_corpus (4.5\%) \\
        
        \textcolor[HTML]{ead9bb}{\rule{8pt}{8pt}} & NuminaMath (1.92\%)
        &  &  \\
        
        \textcolor[HTML]{f0e3cd}{\rule{8pt}{8pt}} & Synthia\_v15 (0.1\%)
        &  &  \\
        \bottomrule
        \end{tabular}
        }
    \end{minipage}
    \caption{Data mixture used in the pre-training stage. Up: Overall composition of the pre-training data samples, including general caption data and other data sources. Bottom: Detailed breakdown of the other data, covering a diverse set of domain-specific datasets.}
    \label{fig:pretrain_data}
\end{figure*}

We construct a diverse data mixture for the pre-training stage, as summarized in Figure~\ref{fig:pretrain_data} provides a visualized overview of the data distribution. Overall, the data mixture in this stage contains approximately 121M samples.

Overall, the majority of our training data consists of general caption data (64\%), which primarily serves to strengthen image–text alignment and to substantially broaden the visual distribution exposed to the model. Beyond this core component, we further enrich the model’s knowledge coverage by incorporating a diverse set of additional data categories, including document, fine-grained, scientific, OCR, code, math, and pure text data.

The general caption data are mainly composed of OpenImages~\citep{openimages}, SA-1B~\citep{kirillov2023segment}, DenseFusion-1M~\citep{li2024densefusion}, VL3-Syn7M~\citep{zhang2025videollama}, PixmoCap~\citep{deitke2025molmo}, Ureader~\citep{Ye2023UReaderUO}, FineVision~\citep{wiedmann2025finevision}, our recaptioned data and the video frame images mentioned in stage 1, providing large-scale and diverse image–caption supervision.

Among the remaining categories, document data account for the largest proportion. This category is particularly important for multi-modal models, as it strongly supports OCR and fine-grained visual recognition, while its typically high image resolution also enhances the model’s ability to process high-resolution visual inputs.

Fine-grained data form the next major component and include grounding and region-caption annotations, which are described in detail in subsequent sections.

We also incorporate pure text data, such as general QA and math QA, where only high-quality filtered text is used. Introducing such text-only data effectively mitigates catastrophic forgetting in the language component during multi-modal training.

In addition, science data enable the model to acquire complex, multidisciplinary knowledge, OCR data further improve the model’s ability to perceive and reason about textual content in visual scenes, and code data help establish a stronger connection between programming capabilities and multi-modal understanding.

\paragraph{Image Region Grounding and Region Caption.}
To equip the vision encoder and VLM with fine-grained, localization-aware perception, we additionally curate two complementary types of region-level image supervision: \emph{image grounding} and \emph{region-centric captioning}. For image grounding, we merge image grounding/detection datasets into a unified grounding corpus containing 7.7 million samples. From this corpus, we construct question--answer pairs that require the model to predict bounding-box coordinates corresponding to textual descriptions of objects or regions. To resolve annotation inconsistencies across heterogeneous grounding datasets, we perform careful format normalization for each dataset individually. All bounding boxes are converted into a unified integer-based coordinate system in the $[0, 1000]$ relative space, where $(0,0)$ denotes the top-left corner and $(1000,1000)$ denotes the bottom-right corner of the image. We adopt an integer coordinate space rather than floating-point representations to better align with LLM tokenization, as predicting discrete integers is empirically more stable and easier for autoregressive language models than regressing high-precision decimal values.

Complementary to grounding, we further self-construct 1.5 million \textbf{region caption QA pairs} to strengthen region-aware semantic understanding. In this setting, questions specify a spatial region using the same normalized $[0,1000]$ coordinate format, and the model is required to describe or reason about the visual content within the queried region. By jointly training on region-to-text (captioning) and text-to-region (grounding) supervision, the model learns a bidirectional alignment between spatial localization and semantic content, substantially improving its region-level perception and reasoning capabilities.

\subsection{Stage 3: Supervised Fine-Tuning}
Following pre-training, we advance to the SFT stage to align the model's multimodal capabilities with user intent.

\subsubsection{Data Mixture}
\begin{figure*}[tbp]
    \centering
    \vspace{-0.2in}
    \caption{Image data composition of the SFT stage.}
    \includegraphics[width=\linewidth]{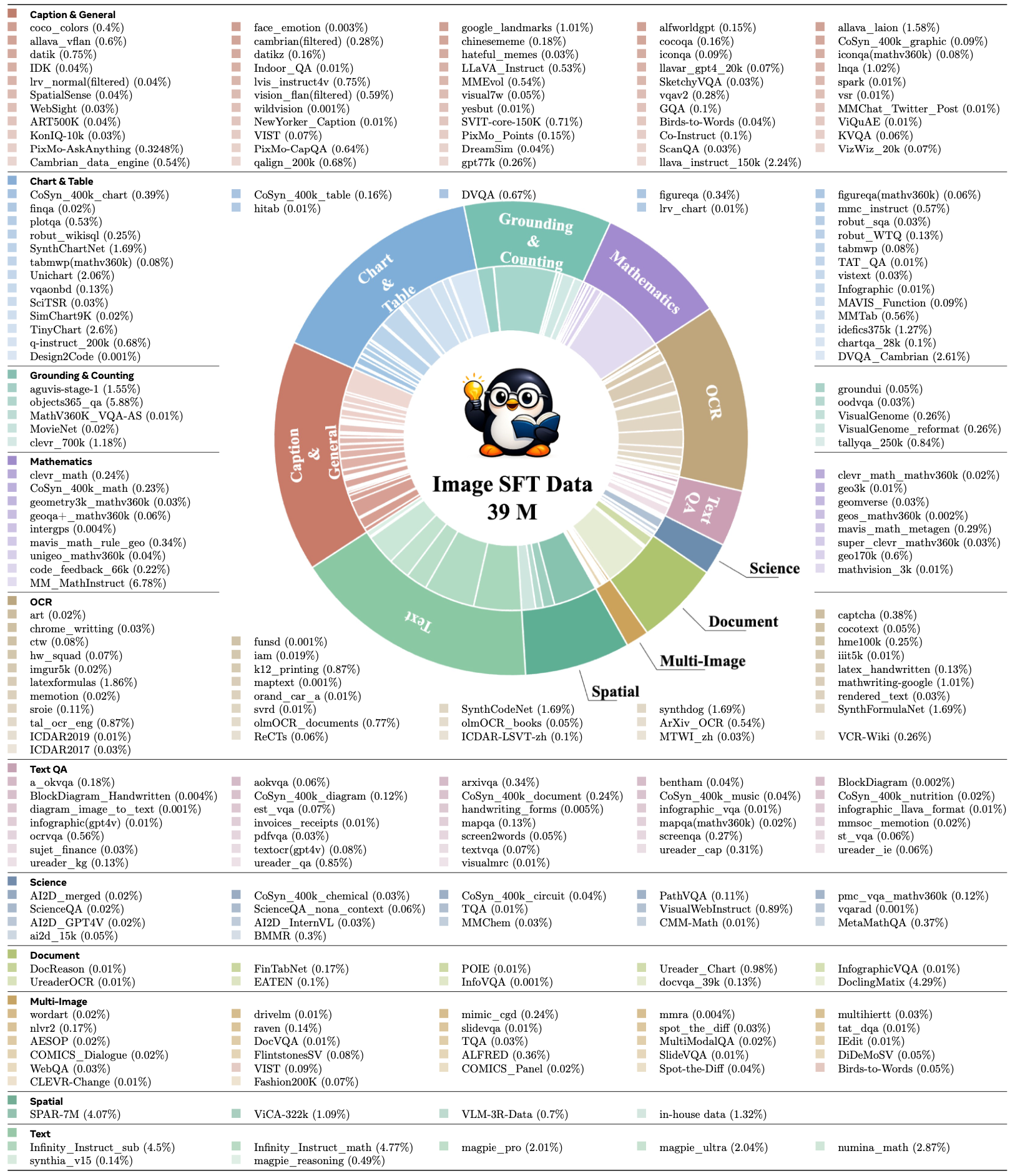}
    \vspace{-0.1in}
    \label{fig:sft_image_mixture}
\end{figure*}

\begin{figure*}[tbp]
    \centering
    \vspace{-0.2in}
    \includegraphics[width=\linewidth]{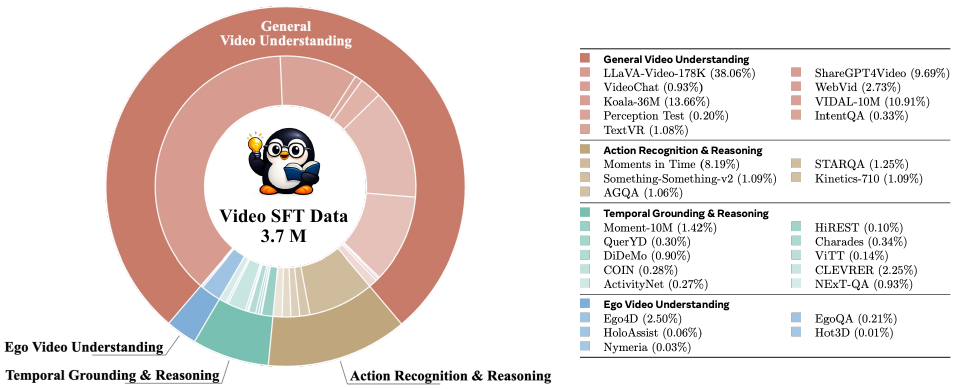}
    \caption{Video data composition of the SFT stage. The datasets are categorized into four domains: General video understanding, action recognition \& reasoning, temporal grounding \& reasoning, and ego video understanding, with specific sampling ratios indicated in parentheses for each data source.}
    \vspace{-0.1in}
    \label{fig:sft_video_mixture}
\end{figure*}

We construct a high-quality SFT dataset designed to cover a broad spectrum of multimodal capabilities, from fundamental perception to complex reasoning. The data compositions for image and video modalities are visualized in Figure~\ref{fig:sft_image_mixture} and Figure~\ref{fig:sft_video_mixture}, respectively.
In both figures, we adopt a nested visualization to illustrate the diversity of our training data. The \textbf{outer circle} represents the high-level semantic domains (e.g., \textit{OCR}, \textit{Science}, \textit{Math}, \textit{Grounding}), defining the core competencies of the model. The \textbf{inner ring} breaks these domains down into fine-grained sub-tasks and specific datasets, illustrating the relative proportion of samples dedicated to each specialized capability.

\paragraph{Image SFT Mixture.} As shown in Figure~\ref{fig:sft_image_mixture}, 
during the SFT stage, we curate a large-scale and highly diverse collection of QA data spanning a wide range of domains and real-world scenarios, with the goal of covering nearly all practical usage cases of the model. After rigorous data cleaning and filtering, we carefully balance the proportion of data across different scenarios to avoid domain bias. This balanced data distribution enables the model to achieve consistently strong performance across multiple sub-domains rather than overfitting to a small set of tasks. Furthermore, we explicitly design the data mixture ratios according to task difficulty, ensuring that more challenging tasks receive sufficient supervision while maintaining overall training stability. The final image training data contains 39M samples.

\begin{itemize}
\item \textbf{General \& Caption, Text (~32.6\%)}. Among all data categories, General \& Caption data, together with pure Text data, play a foundational role. These data types are crucial for strengthening the model’s general comprehension ability, language reasoning capacity, and robustness on open-ended, everyday tasks. They form the backbone of the model’s instruction-following and general-purpose problem-solving capabilities.

\item \textbf{Document, Chart \& Table (~20.9\%)}. In addition, Document as well as Chart \& Table data are critical for fine-grained and structured understanding tasks. Document data enhance the model’s ability to reason over complex layouts, long-context textual structures, and visually rich documents, while Chart and Table data further strengthen numerical reasoning, relational understanding, and structured data interpretation. Incorporating these data significantly improves the model’s performance on detail-oriented tasks such as document QA, table understanding, and chart-based reasoning, which are essential for real-world analytical and professional applications.

\item \textbf{OCR, Text QA (~16.6\%)}. we further incorporate OCR and Text QA data to strengthen the model’s capability in text recognition and language-centric reasoning. These data enable the model to accurately perceive and interpret text embedded in natural scenes and documents, which is essential for tasks such as scene text understanding, document reading, and multi-modal question answering. More importantly, text-oriented supervision encourages the model to perform fine-grained semantic interpretation of visual content, particularly for images containing textual cues and subtle semantic details. By grounding visual understanding in explicit textual reasoning, the model becomes more sensitive to detailed semantics and nuanced visual concepts.

\item \textbf{Grounding \& Counting (~10.1\%)}. To further enhance fine-grained visual reasoning, we include Grounding and Counting data. Grounding data explicitly align textual descriptions or questions with specific visual regions, encouraging the model to learn precise cross-modal correspondence between language and visual entities. Counting data, on the other hand, improve the model’s ability to reason about quantities, object multiplicity, and spatial distribution. Together, these data types significantly strengthen the model’s spatial awareness, object-level reasoning, and robustness on compositional visual tasks.

\item \textbf{Mathematics (~8.9\%)}. We introduce Math data to improve the model’s numerical reasoning and multi-step problem-solving capabilities. Math data expose the model to structured logical reasoning, symbolic manipulation, and arithmetic computation, which are crucial for solving complex quantitative problems. By integrating math-oriented supervision into the SFT stage, the model gains stronger reasoning depth and generalization ability, benefiting not only mathematical tasks but also broader domains that require precise logical inference.

\item \textbf{Multi-image, Science (~3.71\%)}. Finally, we include an Others category that primarily consists of Multi-image and Science data to further broaden the model’s capability boundary. Multi-image data expose the model to inputs containing multiple related images, encouraging it to perform cross-image comparison, temporal or logical association, and joint reasoning over multiple visual contexts. This type of supervision is particularly important for tasks that require relational understanding, consistency reasoning, and multi-view or multi-instance analysis. In addition, Science data introduce domain-specific knowledge and complex reasoning patterns from scientific disciplines such as physics, chemistry, biology, and engineering. These data often involve abstract concepts, structured explanations, and multi-step logical inference, which help enhance the model’s ability to handle knowledge-intensive and analytically demanding tasks. Incorporating science-oriented supervision further improves the model’s generalization to professional and educational scenarios, complementing the more general-purpose data categories in the SFT stage.
\end{itemize}

\paragraph{Video SFT Mixture.}
As illustrated in Figure~\ref{fig:sft_video_mixture}, our video instruction tuning dataset is constructed to balance holistic video comprehension with fine-grained temporal and agent-centric capabilities. 
The mixture comprises four distinct pillars:
\begin{itemize}
    \item \textbf{General Video Understanding (77.6\%):}
    Serving as the cornerstone of our corpus, this category integrates massive-scale datasets such as LLaVA-Video-178K~\cite{zhang2024llava} and VIDAL-10M~\cite{zhulanguagebind}. 
    It ensures the model's proficiency in fundamental tasks, including video captioning, open-ended question answering, and long-context summarization across diverse video domains.

    \item \textbf{Action Recognition and Reasoning (12.7\%):}
    To enhance the model's sensitivity to dynamic events, we incorporate specialized datasets like Moments in Time~\cite{monfort2019moments} and Kinetics-710~\cite{li2022uniformerv2}. 
    This data focuses on empowering the model to identify complex human actions and reason about action semantics and contextual relationships within evolving scenes.

    \item \textbf{Temporal Grounding and Reasoning (6.9\%):}
    We prioritize temporal precision by including datasets such as Moment-10M~\cite{qian2024momentor} and CLEVRER~\cite{Yi*2020CLEVRER:}. 
    This trains the model not only to identify \textit{what} implies an event but to localize and reason about \textit{when} it occurs, including cases with explicit start and end timestamps and structured temporal dependencies.

    \item \textbf{Ego Video Understanding (2.8\%):}
    Recognizing the importance of embodied perception and first-person activity understanding, we introduce a dedicated subset featuring Ego4D~\cite{grauman2022ego4d} and EgoQA~\cite{nguyen2024encoding}. 
    This data equips the model with the ability to interpret visual information from an egocentric viewpoint, crucial for tasks involving human-object interaction and navigation.
\end{itemize}

\section{Experiment}
\label{sec:exp}

\subsection{Implementation Details}
We detail the training configuration and implementation choices adopted at each stage of the \modelname framework. Across all stages, we employ a cosine learning rate decay schedule with a warm-up ratio of $3\%$ of the total training steps. The maximum sequence length is set to $16{,}384$ tokens, among which up to $10{,}240$ tokens are allocated to visual inputs.

During the vision encoder training stage, scale-specific initialization strategies are applied. The vision encoder is initialized from Qwen3-0.6B, while the language backbone is initialized from Qwen3-1.7B and Qwen3-8B for the corresponding model scales. Cross-modal feature alignment is achieved through a lightweight projection module implemented as a two-layer MLP with GELU activation.
 
During this stage, we optimize only the vision encoder and the projection module. In the initial phase, both components use a learning rate of $1.0 \times 10^{-3}$, with VL3-SigLIP-NaViT~\cite{zhang2025videollama} serving as the feature reconstruction target. For the second phase, we remove the feature reconstruction loss and decrease the vision encoder's learning rate to $5.0 \times 10^{-4}$, keeping the projector's rate at $1.0 \times 10^{-3}$."
During the pre-training stage, the vision encoder learning rate is further decreased to $1.0 \times 10^{-4}$. In the supervised fine-tuning stage, all model components are jointly optimized with a unified learning rate of $1.0 \times 10^{-5}$. 

To improve computational efficiency when processing video inputs, visual tokens extracted from videos are spatially downsampled by a factor of 2 using bilinear interpolation immediately after the vision encoder.

For video preprocessing, frames are initially extracted at a fixed rate of one frame per second using FFmpeg. If the resulting frame sequence exceeds a predefined limit, uniform temporal subsampling is applied. The total number of frames ($max\_frames$) is capped at $180$, which is sufficient to cover the temporal span of most videos shorter than three minutes. Finally, the Priority-Aware Visual Token Encoding and Compression method is applied to further compress the visual tokens.

\begin{center}
\renewcommand{\arraystretch}{1.5}
\setlength{\tabcolsep}{8pt}
\setcounter{hcellcount}{0}
\begin{table}[t!]
\caption{
Results comparison for 2B model variants. Best-performing results among all compared models are shown in \textbf{bold}, while the second-best results are \underline{underlined}.
}
\label{tab:2b_results}
\vspace{0.1in}
\centering
\begin{tabular}{ll 
>{\centering\arraybackslash}p{1.8cm} c c c c c}
\toprule
& & \shortstack{\textbf{\modelname-VL} \\ 2B} 
& \shortstack{\textbf{Qwen3-VL} \\ 2B \citep{Bai2025Qwen3VLTR}} 
& \shortstack{\textbf{InternVL3.5} \\ 2B \citep{wang2025internvl3}} 
& \shortstack{\textbf{Gemma3n} \\ E2B-it \citep{Deepmind}} 
& \shortstack{\textbf{SmolVLM2} \\ 2.2B \citep{marafioti2025smolvlm}} \\
\midrule

\multirow{5}{*}{\shortstack[l]{\textbf{Chart}\\ \textbf{/OCR}\\ \textbf{/Doc}}}
& InfoVQA           & \hc{\textbf{77.8}} & \underline{72.4} & 70.8 & 51.9 & 43.0 \\
& ChartQA           & \hc{\textbf{86.6}} & 76.9  & \underline{80.7} & 65.8 & 68.7 \\
& DocVQA            & \hc{\textbf{94.1}} & \underline{93.3} & 89.4 & 78.4  & 80.0 \\
& CharXiv$_{DQ/RQ}$   & \hc{\textbf{66.4/35.8}} & 62.3 /26.8 & \underline{65.0/31.6} & 60.1/27.0 & 36.9/15.5 \\
& OCRBench          & \hc{810} & \textbf{858}  & \underline{836} & 700 & 729 \\
\arrayrulecolor{black!10}\midrule
\arrayrulecolor{black!10}

\multirow{5}{*}{\shortstack[l]{\textbf{General}\\ \textbf{Knowledge}\\ \textbf{/Multi-}\\ \textbf{-Image}}}
& AI2D              & \hc{\textbf{80.7}} & 76.9  & \underline{78.8} & 74.6 & 70.0 \\
& RealWorldQA       & \hc{\textbf{70.2}} & \underline{63.9}  & 62.0 & 59.9 & 58.3 \\
& V-star            & \hc{\textbf{83.8}} & \underline{74.9} & 69.1 & 46.0 & 51.8 \\
& MMMU-Pro          & \hc{31.4} & \textbf{36.5} & \underline{31.6} & 28.0 & 20.1 \\
& BLINK             & \hc{\underline{51.7}} & \textbf{53.8}  & 36.6 & 44.1 & 44.0 \\
\midrule

\multirow{3}{*}{\textbf{Math}}
& MathVista         & \hc{\textbf{67.3}} & \underline{61.3} & 60.8 & 50.4 & 51.5 \\
& MathVerse         & \hc{35.9} & \textbf{52.1}  & \underline{39.6} & 22.5 & 21.5 \\
& LogicVista        & \hc{\underline{41.3}} & 35.8 & \textbf{47.7} & 33.9 & 24.8 \\
\midrule

\multirow{7}{*}{\textbf{Video}}
& MVBench           & \hc{\underline{65.5}} & 61.7  & \textbf{65.9} & 46.8 & 46.3 \\
& LongVideoBench    & \hc{\textbf{59.5}} & 52.1 & \underline{57.4} & 43.0 & 49.7 \\
& VideoMME          & \hc{57.4} & \textbf{61.9}  & \underline{58.4} & 47.0 & 52.1 \\
& Egochema          & \hc{\textbf{57.6}} & \underline{55.7}  & 50.5 & 48.0 & 34 \\
& MMVU              & \hc{\textbf{42.7}}  & \underline{41.7}  & \underline{42.7} & 34.5 & 33.5 \\
& CharadesSTA       & \hc{\textbf{56.2}} & \underline{54.5}  & 21.9 & 5.5 & 9.5 \\
& NextQA            & \hc{\textbf{79.9}} & \underline{76.9}  & 76.1 & 65.4 & 62.4 \\
& ActivityNetQA     & \hc{\textbf{61.5}} & \underline{59.7} & 58.3 & 51.5 & 52.6 \\
& Perception Test   & \hc{\textbf{70.4}} & 64.5 & \underline{64.7} & 48.6 & 51.6 \\
\arrayrulecolor{black}
\bottomrule
\end{tabular}

\begin{tikzpicture}[remember picture, overlay]
  \coordinate (start) at (hc-1.north west);
  \coordinate (end)   at (hc-\thehcellcount.south east);
  
  \path ([shift={(-1.8em,  4.5em)}]start) coordinate (boxNW);
  \path ([shift={( 1.8em, -1.5em)}]end) coordinate (boxSE);
  \path (boxNW |- boxSE) coordinate (boxSW);
  \path (boxSE |- boxNW) coordinate (boxNE);
  
  \begin{pgfonlayer}{floatbox}
    \node[
      draw=black!20,
      line width=1.5pt,
      fill=white,
      rounded corners=12pt,
      blur shadow={shadow blur steps=15, shadow xshift=2pt, shadow yshift=-2pt, shadow opacity=40},
      inner sep=0pt,
      fit={(boxNW) (boxSE)}
    ] (box) {};
  \end{pgfonlayer}

  \begin{pgfonlayer}{foreground}
    \foreach \i in {1,...,\thehcellcount} {
      \node[inner sep=0pt, outer sep=0pt, anchor=base, font=\fontsize{10pt}{13pt}\fontseries{m}\selectfont]
        at (hc-\i.base) {\hctext{\i}};
    }
    
    \node[anchor=south, font=\fontsize{11pt}{11pt}\selectfont] at ([yshift=1.0em]hc-1.north) {\shortstack{\textbf{\modelname-VL} \\ 2B}};
    
    \draw[black, line width=0.5pt] 
      ([yshift=0.94em]hc-1.north west -| boxNW) -- ([yshift=0.94em]hc-1.north east -| boxNE);
    
    \draw[black!10, line width=0.4pt] 
      ([yshift=-0.7em]hc-5.south west -| boxNW) -- ([yshift=-0.7em]hc-5.south east -| boxNE);
    
    \draw[black!10, line width=0.4pt] 
          ([yshift=-0.5em]hc-10.south west -| boxNW) -- ([yshift=-0.5em]hc-10.south east -| boxNE);

    \draw[black!10, line width=0.4pt] 
      ([yshift=-0.5em]hc-13.south west -| boxNW) -- ([yshift=-0.5em]hc-13.south east -| boxNE);
    
  \end{pgfonlayer}
\end{tikzpicture}
\end{table}
\end{center}

\begin{center}
\renewcommand{\arraystretch}{1.5}
\setlength{\tabcolsep}{8pt}
\setcounter{hcellcount}{0}
\begin{table}[t!]
\caption{
Results comparison for 8B model variants. Best-performing results among all compared models are shown in \textbf{bold}, while the second-best results are \underline{underlined}.
}
\label{tab:8b_results}
\vspace{0.1in}
\centering
\begin{tabular}{ll 
>{\centering\arraybackslash}p{1.8cm} c c c c}
\toprule
& & \shortstack{\textbf{\modelname-VL} \\ 8B} 
& \shortstack{\textbf{Qwen3-VL} \\ 8B \citep{Bai2025Qwen3VLTR}} 
& \shortstack{\textbf{InternVL-3.5} \\ 8B \citep{wang2025internvl3}}  
& \shortstack{\textbf{OpenAI GPT-5} \\ nano~\cite{singh2025openai}} \\
\midrule

\multirow{5}{*}{\shortstack[l]{\textbf{Chart}\\ \textbf{/OCR}\\ \textbf{/Doc}}}
& InfoVQA           & \hc{\textbf{86.8}} & \underline{83.1} & 79.1 & 49.2 \\
& ChartQA           & \hc{\textbf{90.5}} & \underline{89.6} & 86.7 & 48.6 \\
& DocVQA            & \hc{\textbf{96.2}} & \underline{96.1} & 92.3 & 78.3 \\
& CharXiv$_{DQ/RQ}$    & \hc{\underline{75.7/40.0}} & \textbf{83.0/46.4} & 72.2/44.4 & 64.4/31.7 \\
& OCRBench          & \hc{\underline{852}} & \textbf{896}  & 840 & 701 \\
\arrayrulecolor{black!10}\midrule
\arrayrulecolor{black!10}

\multirow{5}{*}{\shortstack[l]{\textbf{General}\\ \textbf{Knowledge}\\ \textbf{/Multi-}\\ \textbf{image}}}
& AI2D              & \hc{\textbf{86.1}} & \underline{85.7} & 84.0 & 65.7 \\
& RealWorldQA       & \hc{\textbf{75.8}} & \underline{71.5} & 67.5 & 60.7 \\
& V-star            & \hc{\textbf{90.2}} & \underline{90.1} & 70.7 & 63.4 \\
& MMMU-Pro          & \hc{\underline{40.2}} & \textbf{55.9} & 39.7 & 36.5 \\
& BLINK             & \hc{58.2} & \textbf{69.1} & \underline{59.5} & 42.2 \\
\midrule

\multirow{3}{*}{\textbf{Math}}
& MathVista         & \hc{\textbf{77.4}} & \underline{77.2} & 74.2 & 40.9 \\
& MathVerse         & \hc{50.8} & \textbf{62.1} & \underline{55.8} & 27.0 \\
& LogicVista        & \hc{53.8} & \underline{55.3} & \textbf{57.3} & 40.5 \\
\midrule

\multirow{7}{*}{\textbf{Video}}
& MVBench           & \hc{\underline{71.7}} & 68.7 & \textbf{72.1} & 52.9 \\
& LongVideoBench    & \hc{\textbf{67.0}} & \underline{62.6} & 62.1 & 38.1 \\
& VideoMME          & \hc{\underline{66.2}} & \textbf{71.4} & 66.0 & 49.4 \\
& Egochema          & \hc{\underline{67.0}} & \textbf{70.2} & 61 & 34.8 \\
& MMVU              & \hc{\underline{53.9}}  & \textbf{58.7}  & 51.5 & 51.0 \\
& CharadesSTA       & \hc{\textbf{61.4}} & \underline{56.0} & 32.8 & 5.0 \\
& NextQA            & \hc{\textbf{85.4}} & 82.3  & 81.3 & 59.3 \\
& ActivityNetQA     & \hc{\textbf{65.2}} & \underline{63.7} & 60.1 & - \\
& Perception Test   & \hc{\textbf{78.0}} & 72.7 & 72.7 & - \\
\arrayrulecolor{black}
\bottomrule
\end{tabular}

\begin{tikzpicture}[remember picture, overlay]
  \coordinate (start) at (hc-1.north west);
  \coordinate (end)   at (hc-\thehcellcount.south east);
   
  \path ([shift={(-1.8em,  4.5em)}]start) coordinate (boxNW);
  \path ([shift={( 1.8em, -1.5em)}]end) coordinate (boxSE);
  \path (boxNW |- boxSE) coordinate (boxSW);
  \path (boxSE |- boxNW) coordinate (boxNE);
   
  \begin{pgfonlayer}{floatbox}
    \node[
      draw=black!20,
      line width=1.5pt,
      fill=white,
      rounded corners=12pt,
      blur shadow={shadow blur steps=15, shadow xshift=2pt, shadow yshift=-2pt, shadow opacity=40},
      inner sep=0pt,
      fit={(boxNW) (boxSE)}
    ] (box) {};
  \end{pgfonlayer}

  \begin{pgfonlayer}{foreground}
    \foreach \i in {1,...,\thehcellcount} {
      \node[inner sep=0pt, outer sep=0pt, anchor=base, font=\fontsize{11pt}{11pt}\fontseries{m}\selectfont]
        at (hc-\i.base) {\hctext{\i}};
    }
     
    \node[anchor=south, font=\fontsize{11pt}{11pt}\selectfont] at ([yshift=1.0em]hc-1.north) {\shortstack{\textbf{\modelname-VL} \\ 8B}};
     
    \draw[black, line width=0.5pt] 
      ([yshift=0.94em]hc-1.north west -| boxNW) -- ([yshift=0.94em]hc-1.north east -| boxNE);
     
    
    \draw[black!10, line width=0.4pt] 
      ([yshift=-0.5em]hc-5.south west -| boxNW) -- ([yshift=-0.5em]hc-5.south east -| boxNE);
     
    \draw[black!10, line width=0.4pt] 
          ([yshift=-0.7em]hc-10.south west -| boxNW) -- ([yshift=-0.7em]hc-10.south east -| boxNE);

    \draw[black!10, line width=0.4pt] 
      ([yshift=-0.7em]hc-13.south west -| boxNW) -- ([yshift=-0.7em]hc-13.south east -| boxNE);
     
  \end{pgfonlayer}
\end{tikzpicture}
\end{table}
\end{center}
\paragraph{Baselines}
To comprehensively evaluate the image understanding performance of \modelname, we compare it against a diverse set of baseline models. For the 2B configuration, we select several widely adopted models designed for on-device and efficient multimodal understanding, including Gemma3n-E2B-it~\citep{Deepmind}, SmolVLM2~\citep{marafioti2025smolvlm}, InternVL3.5-2B~\citep{wang2025internvl3}, and Qwen3-VL-2B~\citep{Bai2025Qwen3VLTR}. For the 8B configuration, we evaluate the corresponding 8B variants of these models (if available) and additionally include the close-source model GPT-5-nano~\citep{openai2025gpt5nano} for comparison.

\subsection{Inference Settings}
\paragraph{Image} To ensure robustness and reproducibility, we employ a deterministic inference strategy for \modelname. Specifically, decoding is performed using greedy or near-greedy sampling, with the temperature set to $0.0$ or $0.1$, while $top_p$ and $top_k$ are fixed to $1.0$ and $50$, respectively. The maximum generation length is constrained to match the configuration used during training, ensuring consistent model behavior across training and inference. 

\paragraph{Video} 
Building upon the deterministic inference settings established for images, we introduce additional configurations for video inputs: specifically, we cap the maximum number of frames at $300$ and set the frame rate up to $3$ FPS. Two frame sampling strategies are considered:
\begin{itemize}
    \item 1) TRA, follows the same sampling procedure used during training (see Section~\ref{sec:pac}).
    \item 2) TRA-codec, leverages I-frame (key frames encoded in video file) information from the encoded video. In this strategy, key frames $T_k$ are selected from I-frames, while intermediate frames $T_i$ are sampled according to the target FPS within each I-frame interval. When the number of I-frames exceeds the maximum frame budget, I-frames are uniformly subsampled to form $T_k$, and longer I-frame intervals are prioritized when sampling $T_i$.
\end{itemize}
The results reported in Table~\ref{tab:2b_results} correspond to the best-performing configuration selected from all the settings and sampling strategies described above.

To facilitate fair and standardized evaluation, all baseline results are obtained using the \texttt{lmms-eval} toolkit~\cite{lmms_eval2024}. For models not natively supported by \texttt{lmms-eval}, such as {Gemma3n-E2B-it, we adapt \modelname’s inference template to standardize prompt formatting and ensure equitable comparisons across architectures. Due to the limited context length of Gemma3n-E2B-it, we cap the input to 32 video frames using uniform (linspace) sampling, which doubles the 16-frame context reported in the Gemma3 technical report~\citep{team2025gemma}.

\subsection{Image Benchmarks}

To rigorously assess Penguin-VL's capabilities in static image recognition and complex visual reasoning, we conduct an extensive evaluation over a curated suite of representative benchmarks. These benchmarks are chosen to span diverse visual domains and varying cognitive demands, enabling a holistic characterization of the model’s performance. Concretely, our image evaluation covers three complementary dimensions: fine-grained text and data interpretation in visual contexts, complex mathematical and logical reasoning, and cross-image synthesis grounded in general knowledge.

\paragraph{Document, Chart, and Scene Text Understanding.}
A fundamental requirement for advanced vision--language models is the ability to parse, comprehend, and reason over dense textual and graphical content embedded in images. To assess \modelname’s visual literacy in this setting, we evaluate it on several document- and text-centric benchmarks. \texttt{DocVQA}~\citep{mathew2021docvqa} measures the model’s ability to extract and reason over textual information from scanned documents, placing strong emphasis on accurate OCR and layout understanding. To evaluate data-centric reasoning in graphical formats, we further include \texttt{ChartQA}~\citep{masry2022chartqa} and \texttt{InfoVQA}~\citep{mathew2021docvqa}, which require the interpretation of bar charts, line plots, and complex infographics through the integration of visual cues and numerical reasoning. In addition, scene text understanding is examined using \texttt{OCRBench}~\citep{liu2023hidden}, which focuses on recognizing and comprehending text appearing in unconstrained, cluttered real-world environments.

\paragraph{Mathematical and Logical Reasoning.}
Beyond perceptual accuracy, we investigate Penguin-VL’s capacity for higher-order reasoning that combines visual inputs with quantitative and logical deduction. General visual mathematical problem solving is evaluated using \texttt{MathVista}~\citep{lu2023mathvista}, which include tasks such as geometry reasoning, function interpretation, and visually grounded numerical analysis. To probe more abstract and multi-step reasoning, we further incorporate \texttt{MathVerse}~\citep{zhang2025mathverse} and \texttt{LogicVista}~\citep{xiao2024logicvista}. These benchmarks emphasize visually grounded logical puzzles and complex deduction chains, providing a stricter test of genuine reasoning beyond surface-level pattern recognition.

\paragraph{Multi-image and General Knowledge Reasoning.}
Real-world visual reasoning often extends beyond single static images, requiring the integration of information across multiple views and diverse knowledge domains. To assess these capabilities, we evaluate Penguin-VL on a range of benchmarks targeting cross-image understanding, general knowledge, and scientific reasoning. \texttt{MMMU-Pro}~\citep{yue2024mmmupro} measures expert-level reasoning in professional and academic contexts, frequently demanding the synthesis of general knowledge and information distributed across multiple figures or slides. Similarly, \texttt{BLINK}~\citep{fu2025blink} focuses on multi-image and sequential reasoning while also testing broad general knowledge through change detection, relational inference, and narrative understanding across image sequences. To examine robustness in everyday visual reasoning, \texttt{RealWorldQA}~\citep{realworldqa} evaluates spatial, physical, and common-sense reasoning in realistic scenarios. Scientific diagram understanding is assessed using \texttt{AI2D}~\citep{kembhavi2016diagram}, which requires structured reasoning over textbook-style diagrams with arrows, labels, and schematic layouts. Finally, fine-grained visual perception is tested with \texttt{V-star}~\citep{wu2024v}, a high-resolution benchmark that demands precise attention to subtle visual details often lost at standard resolutions.

\subsubsection{Image Results}

\paragraph{2B Model} Table~\ref{tab:2b_results} presents the benchmark results for 2B-scale models along with their corresponding baselines. \modelname-VL achieves leading performance on most Chart and Document understanding benchmarks. Although it underperforms Qwen3-VL and InternVL3.5 on \texttt{OCRBench}, it still surpasses earlier baselines such as Gemma3n-E2B-it and SmolVLM2 by a margin of 81--110 points. On mathematics-oriented tasks, \modelname attains the best performance on \texttt{MathVista}, while Qwen3-VL demonstrates strong results on two of the four evaluated math benchmarks. For \texttt{LogicVista}, \modelname achieves a score of 41.29, outperforming Qwen3-VL but falling short of InternVL3.5. This gap suggests that \modelname may be mildly constrained by comparatively limited math-focused SFT data. Despite this limitation, \modelname demonstrates exceptional competitiveness on general knowledge benchmarks, claiming top rankings on three of five tasks and achieving state-of-the-art results. It substantially outperforms InternVL3.5—particularly with a commanding 15.1-point margin on \texttt{BLINK}—while only trailing Qwen3-VL. Qwen3-VL's sole advantage emerges on \texttt{MMMU-Pro}.

\paragraph{8B Model} Table \ref{tab:8b_results} reveals that \modelname-VL-8B consistently establishes itself as the frontrunner in this weight class. In the domain of document and graphical interpretation, the model demonstrates a remarkable command over dense visual information, achieving a score of 96.2 on \texttt{DocVQA} and 90.5 on \texttt{ChartQA}. This performance suggests a highly refined ability to handle complex layouts and numerical data within bar charts and line plots, where it frequently surpasses Qwen3-VL 8B. While \modelname-8B takes a slight second to Qwen3-VL in \texttt{OCRBench} and the data-heavy \texttt{CharXiv} benchmark, its scores remain significantly ahead of other contemporaries like InternVL-3.5 and the OpenAI GPT-5 nano, highlighting its specialized strength in fine-grained visual literacy. The model’s performance on general knowledge and scientific reasoning benchmarks further underscores its versatility. \modelname-8B secures leading positions on tasks like \texttt{AI2D} and \texttt{V-star}, indicating a superior capacity for structured reasoning over textbook-style diagrams and high-resolution visual details. Interestingly, while it maintains a lead in foundational math tasks through its 77.4 score on \texttt{MathVista}, it faces stiffer competition in abstract logic and multi-step deduction. On \texttt{MathVerse} and \texttt{LogicVista}, \modelname-8B is outperformed by Qwen3-VL and InternVL-3.5, which suggests that while the model has excellent perceptual grounding for math, it may still be refining the deeper logical chains required for advanced visual puzzles.

\subsection{Video Benchmarks}

To evaluate dynamic visual perception, we assess \modelname-VL’s video understanding capabilities across a diverse set of established benchmarks. These tasks are designed to measure comprehensive temporal and spatial reasoning, spanning short-form question answering, long-form comprehension, and dedicated temporal awareness.

\paragraph{General Video Understanding.}
General short-form video understanding is assessed using multi-choice video question answering (MC-VQA) benchmarks, including \texttt{MVBench}~\citep{li2023mvbench}, \texttt{VideoMME}~\citep{fu2024video}, \texttt{EgoSchema}~\citep{mangalam2024egoschema} and \texttt{PerceptionTest}~\citep{patraucean2024perception}. For open-ended question answering (OE-VQA), we consider \texttt{ActivityNetQA}~\citep{yu2019activitynet}. In addition, we evaluate on \texttt{MMVU}~\citep{zhao2025mmvu}, which combines both MC-VQA and OE-VQA tasks, providing a more comprehensive assessment of baseline video understanding ability.

\paragraph{Long-form Comprehension and Temporal Reasoning.}
To examine \modelname-VL’s capacity for long-form video comprehension, we conduct evaluations on two long-video understanding benchmarks. \texttt{LongVideoBench}~\citep{wu2024longvideobench} focuses on video reasoning over extended video–language interleaved inputs. Finally, to assess temporal awareness and reasoning, we evaluate \modelname on temporal perception and reasoning benchmarks, including \texttt{NextQA}~\citep{xiao2021next} as well as the temporal sentence grounding task on \texttt{Charades-STA}~\citep{gao2017tall}.

\subsubsection{Video Results}
\paragraph{2B Model} Table~\ref{tab:2b_results} also presents a comprehensive evaluation of \modelname-VL across diverse video understanding benchmarks, comparing its performance against the state-of-the-art baselines. In the realm of general video understanding, \modelname-VL demonstrates robust reasoning capabilities, achieving the top performance on \texttt{EgoSchema} with 57.6, \texttt{ActivityNetQA} with 61.5, and the \texttt{Perception Test} with 70.4. It also ties for first place on the comprehensive \texttt{MMVU} benchmark at 42.7, while maintaining competitive scores of 65.5 on \texttt{MVBench} and 57.4 on \texttt{VideoMME}. Regarding long-form video comprehension, \modelname-VL excels on \texttt{LongVideoBench} with a score of 59.5, surpassing the strong Qwen3-VL baseline by 7.4 points. Furthermore, \modelname-VL exhibits superior temporal awareness and reasoning, securing state-of-the-art results on both \texttt{NextQA} with 79.9 and the temporal sentence grounding task \texttt{Charades-STA} with 56.2. The advantage is particularly evident on \texttt{Charades-STA}, where \modelname-VL outperforms InternVL3.5 by a substantial margin of over 34 points, highlighting its precise temporal localization ability. Overall, \modelname-VL establishes itself as a versatile model, effectively balancing short-form reasoning with specialized proficiency in long-context and temporal tasks.

\paragraph{8B Model} Video understanding is the most impressive arena for \modelname-VL-8B, where it claims the top spot in the majority of evaluated metrics. According to Table~\ref{tab:8b_results}, its performance on \texttt{LongVideoBench} and \texttt{NextQA}, reaching 67.0 and 85.4 respectively, showcases an exceptional ability to maintain coherence and temporal awareness across extended sequences. In general video understanding, \modelname-VL-8B secures state-of-the-art results on \texttt{ActivityNetQA} with 65.2 and the \texttt{Perception Test} with 78.0. It also demonstrates highly competitive performance by achieving the second-best scores on \texttt{MVBench} at 71.7, trailing InternVL-3.5 by just 0.4 points, as well as 66.2 on \texttt{VideoMME}, 67.0 on \texttt{EgoSchema}, and 53.9 on \texttt{MMVU}, largely remaining closely behind the Qwen3-VL baseline. Furthermore, its proficiency in precise temporal grounding is evident on \texttt{Charades-STA}, where it achieves a leading score of 61.4, outperforming Qwen3-VL by 5.4 points and surpassing InternVL-3.5 by a substantial 28.6 points. Overall, these results underscore \modelname-VL-8B's robust and comprehensive spatial-temporal reasoning capabilities across various video evaluation frameworks.

\subsection{Ablation Study}
\label{sec:ablation}
\setlength{\aboverulesep}{0pt}
\setlength{\belowrulesep}{0pt}

\definecolor{cBlue}{HTML}{DBEAFE}   
\definecolor{cPurple}{HTML}{F3E8FF} 
\definecolor{cGold}{HTML}{FFE699}   
\definecolor{cOrange}{HTML}{F8CBAD} 
\definecolor{cGray}{HTML}{F3F4F6}   

\newcommand{\cb}{\cellcolor{cBlue}}    
\newcommand{\cp}{\cellcolor{cPurple}}  
\newcommand{\cg}{\cellcolor{cGray}}    
\newcommand{\cgold}{\cellcolor{cGold}} 
\newcommand{\co}{\cellcolor{cOrange}}  

\begin{table}[htb]
    \centering
    \caption{\textbf{Ablation of Penguin-Encoder and comparison for LMM integration.} Results are reported on five benchmarks with the average score. We categorize the training pipelines into four stages, and list approximate data sample scales for each stage. Qwen3VL-ViT and VL3-NaViT are obtained via continued pre-training on large-scale contrastively pre-trained SigLIP models, with Stage~1 data scales taken from their original papers. SigLIP2 (original resolution) follows the official preprocessing, while SigLIP2 (any resolution) adopts NaViT-style dynamic resolution with relative positional encoding. All SigLIP2 variants use siglip2-so400m-patch16-naflex. \textbf{*:}This value is our estimate based on the Qwen3-VL report, which only provides the total number of tokens.}
    \vspace{0.5cm}
    \renewcommand{\arraystretch}{1.6} 
    \resizebox{\textwidth}{!}{
    \begin{tabular}{l ccccc @{\hspace{2em}} cccccc}
        \toprule
        & \multicolumn{5}{c}{\textbf{Training Stages and Configuration}} & \multicolumn{6}{c}{\textbf{Evaluation Results}} \\
        \cmidrule(lr){2-6} \cmidrule(l){7-12}
        
        \textbf{Vision Encoder} &
        \rotatebox{90}{\makecell[l]{\textbf{Stage 0}\\\textbf{Contrastive}\\\textbf{Learning}}} &
        \rotatebox{90}{\makecell[l]{\textbf{Stage 1}\\\textbf{Encoder}\\\textbf{Pretraining}}} &
        \rotatebox{90}{\makecell[l]{\textbf{Stage 2}\\\textbf{Modality}\\\textbf{Alignment}}} &
        \rotatebox{90}{\makecell[l]{\textbf{Stage 3}\\\textbf{SFT}}} &
        \rotatebox{90}{\makecell[l]{\textbf{Total data}\\\textbf{size}}} &
        \rotatebox{90}{\textbf{Avg}} &
        \rotatebox{90}{\textbf{AI2D}} &
        \rotatebox{90}{\textbf{MathVista}} &
        \rotatebox{90}{\textbf{ChartQA}} &
        \rotatebox{90}{\textbf{MMMU-Pro}} &
        \rotatebox{90}{\makecell[l]{\textbf{Realworld}\\\textbf{QA}}} \\
        \midrule
        
        \rowcolor{gray!15} \multicolumn{12}{c}{\textbf{Penguin-Encoder Pretraining Ablation}} \\
        \midrule

         \modelname-Encoder (\textbf{using} random init) & 
        \cp & 
        \cgold & 
        \cg & 
        \cg & 
        \cp & 
        31.3 & 57.2 & 22.0 & 12.4 & 18.8 & 46.0 \\
        \arrayrulecolor{black}\cline{1-1}\arrayrulecolor{cPurple}\cline{2-2}\arrayrulecolor{cGray}\cline{3-5}\arrayrulecolor{cPurple}\cline{6-6}\arrayrulecolor{black}\cline{7-12}
        
        \modelname-Encoder (\textbf{w/o} reconstruction) & 
        \cp & 
        \cgold & 
        \cg & 
        \cg & 
        \cp & 
        32.6 & 55.6 & 29.9 & 11.6 & 18.9 & 46.9 \\
        \arrayrulecolor{black}\cline{1-1}\arrayrulecolor{cPurple}\cline{2-2}\arrayrulecolor{cGray}\cline{3-5}\arrayrulecolor{cPurple}\cline{6-6}\arrayrulecolor{black}\cline{7-12}
        
        \modelname-Encoder (\textbf{w/o} relation loss) & 
        \cp & 
        \cgold & 
        \cg & 
        \cg & 
        \cp & 
        33.3 & 56.3 & 29.5 & 17.4 & 18.2 & 45.0 \\
        \arrayrulecolor{black}\cline{1-1}\arrayrulecolor{cPurple}\cline{2-2}\arrayrulecolor{cGray}\cline{3-5}\arrayrulecolor{cPurple}\cline{6-6}\arrayrulecolor{black}\cline{7-12}
        
        \textbf{\modelname-Encoder} & 
        \cp \multirow{-4}{*}{-} & 
        \cgold \multirow{-4}{*}{\makecell{24M}} & 
        \cg \multirow{-4}{*}{\makecell{DenseFusion\\-1M}} & 
        \cg \multirow{-4}{*}{\makecell{LLaVA-\\665k}} & 
        \cp \multirow{-4}{*}{\(\sim\)24M} & 
        \textbf{34.6} & \textbf{56.4} & \textbf{29.0} & \textbf{17.4} & \textbf{19.3} & \textbf{51.0} \\
        \arrayrulecolor{black}\midrule
        
        \rowcolor{gray!15} \multicolumn{12}{c}{\textbf{LMM Integration Comparison}} \\
        \arrayrulecolor{black}\midrule
        
        VL3-SigLIP-NaViT~\cite{zhang2025videollama} & 
        \cb & 
        \co >37M & 
        \cg & 
        \cg & 
        \cb & 
        43.0 & 60.9 & 32.2 & 40.4 & 21.1 & 60.3 \\
        \arrayrulecolor{black}\cline{1-1}\arrayrulecolor{cBlue}\cline{2-2}\arrayrulecolor{cGold}\cline{3-3}\arrayrulecolor{cGray}\cline{4-5}\arrayrulecolor{cBlue}\cline{6-6}\arrayrulecolor{black}\cline{7-12}
        
        Qwen3VL-32B ViT~\cite{Bai2025Qwen3VLTR} & 
        \cb & 
        \cgold >1B* & 
        \cg & 
        \cg & 
        \cb & 
        47.3 & 64.6 & 34.2 & 53.6 & 22.3 & 62.0 \\
        \arrayrulecolor{black}\cline{1-1}\arrayrulecolor{cBlue}\cline{2-2}\arrayrulecolor{cPurple}\cline{3-3}\arrayrulecolor{cGray}\cline{4-5}\arrayrulecolor{cBlue}\cline{6-6}\arrayrulecolor{black}\cline{7-12}
        
         & 
        \cb & 
        \co - & 
        \cg & 
        \cg & 
        \cb & 
        32.4 & 55.9 & 28.5 & 12.5 & 19.0 & 45.9 \\
        \arrayrulecolor{cBlue}\cline{2-2}\arrayrulecolor{cOrange}\cline{3-3}\arrayrulecolor{cGray}\cline{4-5}\arrayrulecolor{cBlue}\cline{6-6}\arrayrulecolor{black}\cline{7-12}
        
         & 
        \cb & 
        \cgold 24M & 
        \cg & 
        \cg & 
        \cb & 
        36.7 & 59.4 & 30.4 & 19.9 & 21.2 & 52.4 \\
        \arrayrulecolor{cBlue}\cline{2-2}\arrayrulecolor{cGold}\cline{3-3}\arrayrulecolor{cGray}\cline{4-5}\arrayrulecolor{cBlue}\cline{6-6}\arrayrulecolor{black}\cline{7-12}
        
        \multirow{-3}{*}{SigLIP2 (original resolution)~\cite{tschannen2025siglip}} & 
        \cb & 
        \co & 
        \cg & 
        \cg & 
        \cb & 
        39.2 & 61.5 & 32.4 & 25.5 & 22.0 & 54.6 \\
        \arrayrulecolor{black}\cline{1-1}\arrayrulecolor{cBlue}\cline{2-2}\arrayrulecolor{cOrange}\cline{3-3}\arrayrulecolor{cGray}\cline{4-5}\arrayrulecolor{cBlue}\cline{6-6}\arrayrulecolor{black}\cline{7-12}
        
        SigLIP2 (any resolution, 1024 max tokens) & 
        \cb & 
        \co & 
        \cg & 
        \cg & 
        \cb  & 
        45.3 & 65.3 & 33.6 & 46.7 & 22.3 & 58.6 \\
        \arrayrulecolor{black}\cline{1-1}\arrayrulecolor{cBlue}\cline{2-2}\arrayrulecolor{cOrange}\cline{3-3}\arrayrulecolor{cGray}\cline{4-5}\arrayrulecolor{cBlue}\cline{6-6}\arrayrulecolor{black}\cline{7-12}
        
        SigLIP2 (any resolution, 10240 max tokens) & 
        \cb \multirow{-7}{*}{>40B} & 
        \co  & 
        \cg & 
        \cg & 
        \cb \multirow{-7}{*}{>40B} & 
        45.9 & 64.3 & 33.5 & 48.8 & 21.4 & 61.6 \\
        \arrayrulecolor{black}\cline{1-1}\arrayrulecolor{cBlue}\cline{2-2}\arrayrulecolor{cOrange}\cline{3-3}\arrayrulecolor{cGray}\cline{4-5}\arrayrulecolor{cBlue}\cline{6-6}\arrayrulecolor{black}\cline{7-12}
        
        \textbf{\modelname-Encoder} & 
        \cp - & 
        \co \multirow{-4}{*}{240M} & 
        \cg \multirow{-8}{*}{\makecell{DenseFusion\\-1M}} & 
        \cg \multirow{-8}{*}{\makecell{LLaVA-\\665k}} & 
        \cp \(\sim\)240M & 
        \textbf{49.3} & \textbf{65.5} & \textbf{36.3} & \textbf{55.0} & \textbf{24.9} & \textbf{65.0} \\
        
        \arrayrulecolor{black}\bottomrule
    \end{tabular}%
    }
    \label{tab:ablation}
\end{table}

\newcommand{\circnum}[1]{\textcircled{\small #1}}
\paragraph{Initialization and Reconstruction-Loss Ablation.}
We investigate three key design choices of the encoder:
\circnum{1} whether to initialize the Penguin-encoder with pre-trained LLM weights,
\circnum{2} whether to introduce the relation loss, and
\circnum{3} whether to incorporate the full reconstruction loss.
To this end, we adopt a lightweight three-stage training pipeline.
In Stages~1.1 and~1.2, we randomly sample 10\% of the full Stage~1 corpus
(approximately 20M low-resolution and 4M high-resolution samples)
and pre-train only the \emph{encoder} together with the \emph{projector}.
In Stage~2, we perform VLM pre-training on DenseFusion-1M,
with all model parameters remaining trainable.
In Stage~3, we conduct SFT on LLaVA-665k.

After training, we evaluate on five benchmarks:
AI2D, MathVista, ChartQA, MMMU-Pro, and RealworldQA
(Table~\ref{tab:ablation}).
We make two main observations:
\begin{itemize}
    \item \textbf{Initializing from LLM weights consistently strengthens the Penguin-encoder.} 
    Compared to a standard random initialization baseline, which achieves an average score of only 31.3, starting from pretrained LLM weights provides an obvious performance leap, driving the full model's average score to 34.6. This substantial +3.3 absolute improvement highlights the critical advantage of LLM-based initialization. A plausible explanation is that reusing mature LLM weights provides a well-conditioned starting point that allows the encoder to directly inherit the robust sequence modeling capabilities of the pre-trained LLM. This inherent prior makes it significantly easier to learn visual representations that are naturally compatible with the downstream backbone LLM during modality alignment and SFT, while simultaneously ensuring greater training stability.

    \item \textbf{Relation loss is critical to feature reconstruction performance.} 
    While introducing basic reconstruction objectives yields moderate gains over pure LLM initialization (raising the average score in Tab~\ref{tab:ablation} from 32.6 to 33.3), incorporating the relation loss provides a substantial further boost to 34.6. This empirical leap aligns with the inherent nature of attention mechanisms, which prioritize inter-token relationships over absolute individual attributes. By explicitly supervising these inter-patch interactions, the relation loss enables the encoder to more effectively capture the underlying visual space, thereby significantly benefiting downstream multimodal reasoning.
\end{itemize}

\paragraph{Comparisons with Existing Vision Encoders.}
To compare our Penguin-encoder with existing vision encoders,
we conduct LMM integration experiments.
The Penguin-encoder is pre-trained on the full Stage~1 dataset
(approximately 240M samples in total, including 200M low-resolution and 40M high-resolution images)
and compared against representative baselines such as
SigLIP2~\cite{tschannen2025siglip}, VL3-NaViT, and Qwen3-ViT~\cite{Bai2025Qwen3VLTR}.
For fairness, we adopt Qwen3-1.7B as the unified LLM backbone across all settings.
All encoders are connected to the LLM via a randomly initialized MLP projector. For Qwen3-VL ViT, we retain its original merger module in addition to the newly introduced MLP projector.
The integration follows a two-stage procedure:
(i) modality alignment on DenseFusion-1M, and
(ii) SFT on LLaVA-665k.

The results are summarized in Table~\ref{tab:ablation}.
We highlight three main observations:

\begin{itemize}
    \item \textbf{Overall performance advantage.} Penguin-encoder achieves substantially better performance across all benchmarks,
    reaching an average score of 49.3 while using only $\sim$240M pre-training samples.
    We interpret this advantage from a first-principles perspective:
    contrastive encoders (e.g., SigLIP) are optimized for discriminative alignment,
    which is not inherently tailored for reasoning-centric generation.
    In contrast, our \emph{generation-aligned} design directly maps fine-grained visual features
    into the high-level semantic space of the LLM,
    making the representations intrinsically suitable for autoregressive decoding.
    These findings suggest that an architecture initialized from a text-only LLM
    and trained with generative supervision can outperform traditional contrastive-pretrained encoders,
    while avoiding their dependence on massive-scale data (e.g., $>40$B samples)
    and high computational cost.

    \item \textbf{Scalability.}
    With limited data, contrastive encoders may converge faster.
    However, as data scale and task complexity increase,
    Penguin-encoder exhibits progressively larger gains,
    suggesting a higher performance ceiling for LMM integration.

    \item \textbf{Comparison under Matched Data.} Finally, to rule out potential unfair advantages arising from domain overlap
    between our pre-training data and the evaluation benchmarks,
    we conduct a controlled experiment.
    Specifically, we take the pre-trained SigLIP2 encoder
    and further train it using exactly the same data and training recipe
    as the Penguin-encoder (Stage~1).
    Even under matched data and training settings,
    the Penguin-encoder remains clearly superior,
    highlighting the architectural benefit of generation-aligned encoder initialization.
\end{itemize}

\begin{figure*}[tbp] 
\centering
\includegraphics[width=1.\linewidth]{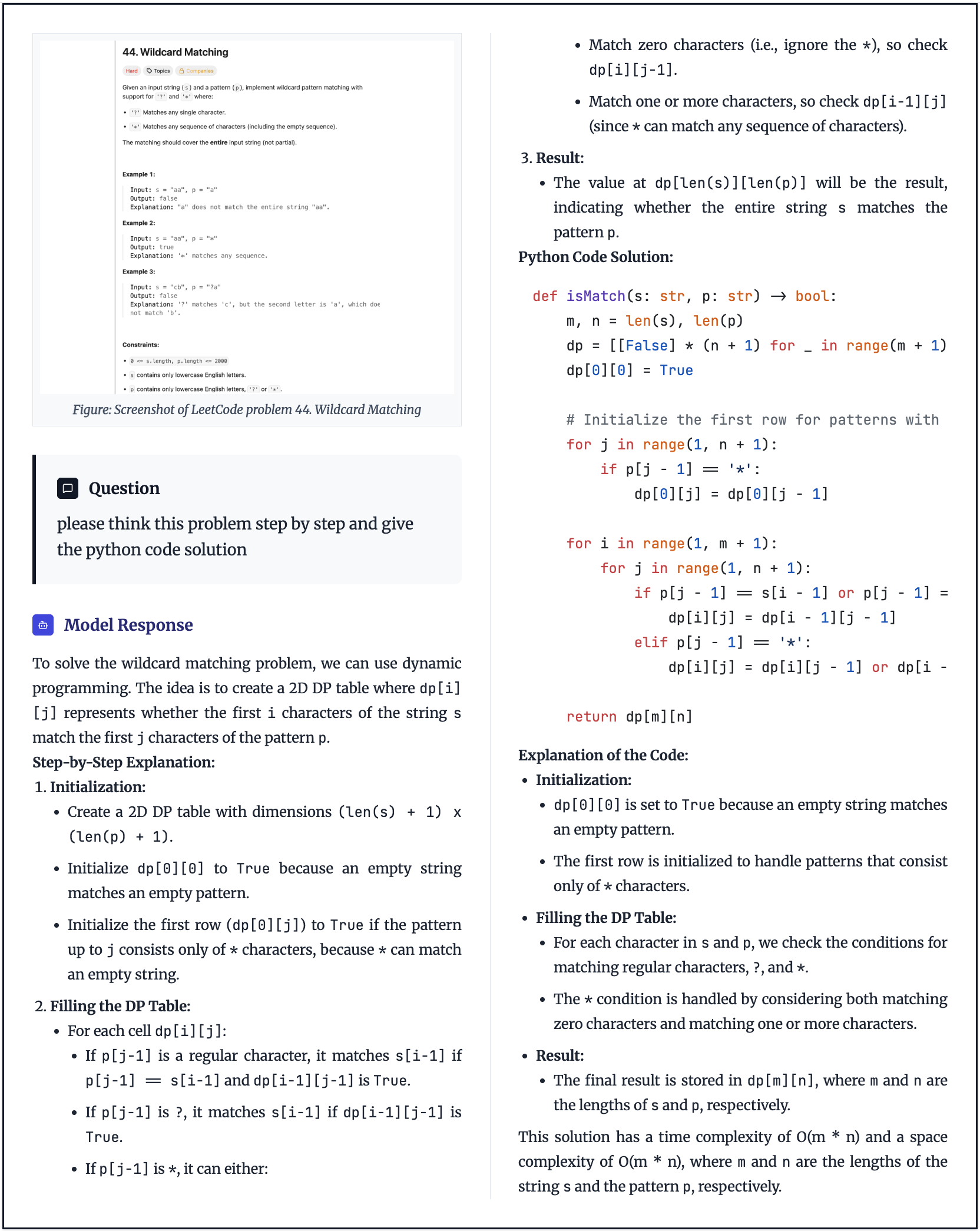} 
\caption{Case study of logical reasoning and algorithm code writing. The input figure caption is for illustration only and is not provided to the model.
}
\label{fig:case_code}
\end{figure*}

\begin{figure*}[tbp] 
\centering
\includegraphics[width=1.\linewidth]{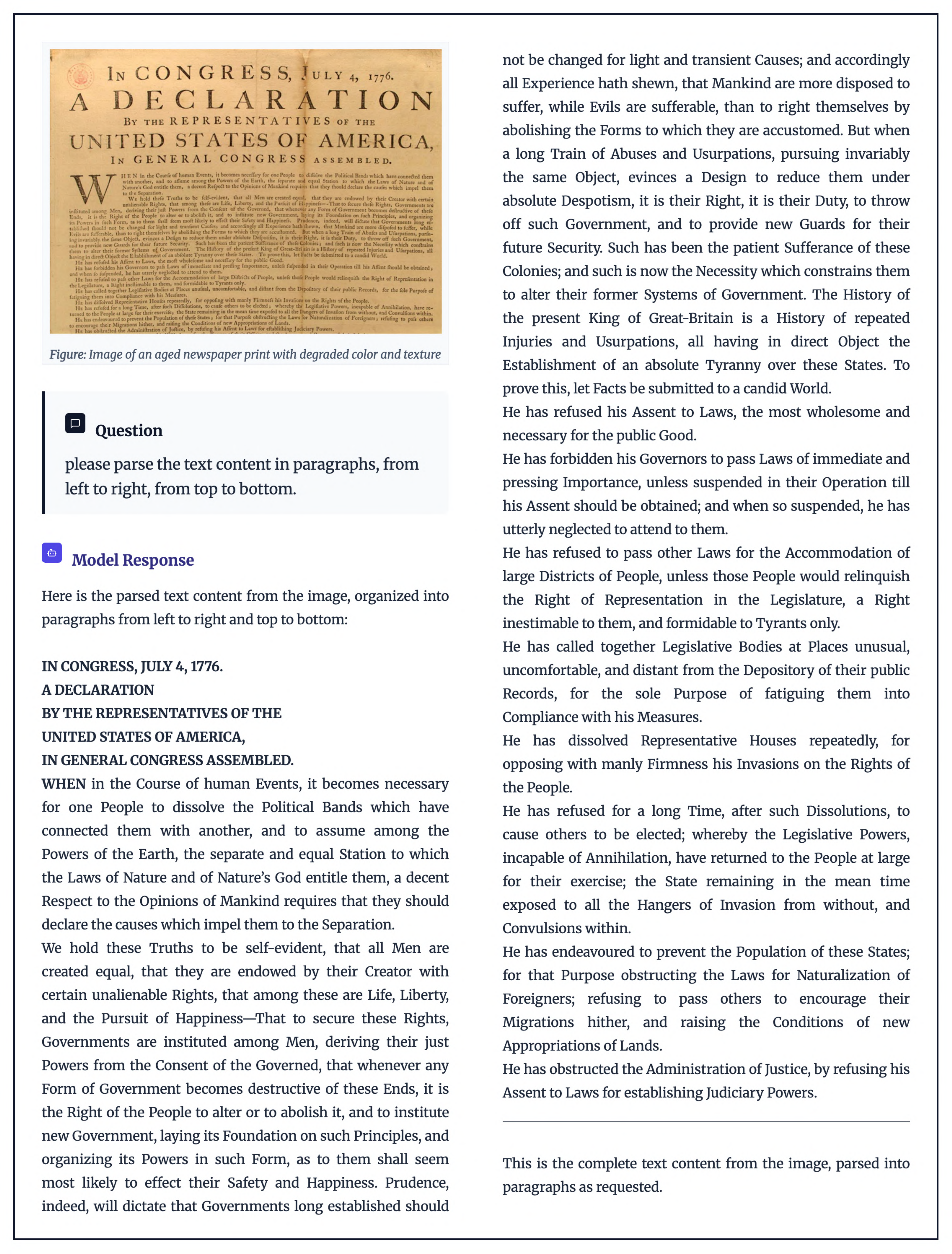} 
\caption{Case study of OCR and dense document understanding. The input figure caption is for illustration only.}
\label{fig:case_ocr}
\end{figure*}

\begin{figure*}[tbp] 
\centering
\includegraphics[width=1.\linewidth]{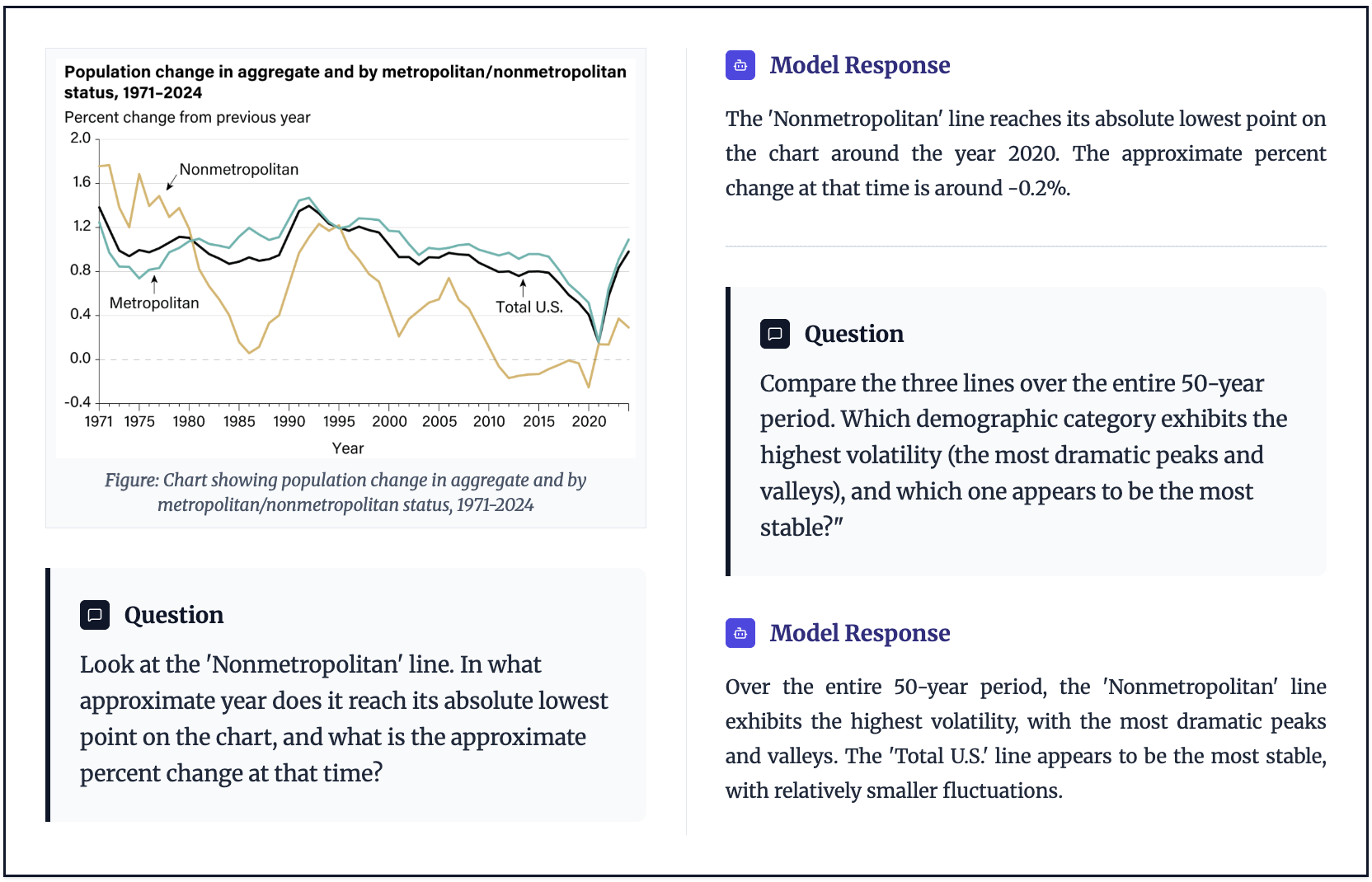} 
\caption{Case study of chart interpretation and understanding. The input figure caption is for illustration only and is not provided to the model.
}
\label{fig:case_chart}
\end{figure*}

\begin{figure*}[tbp] 
\centering
\includegraphics[width=1.\linewidth]{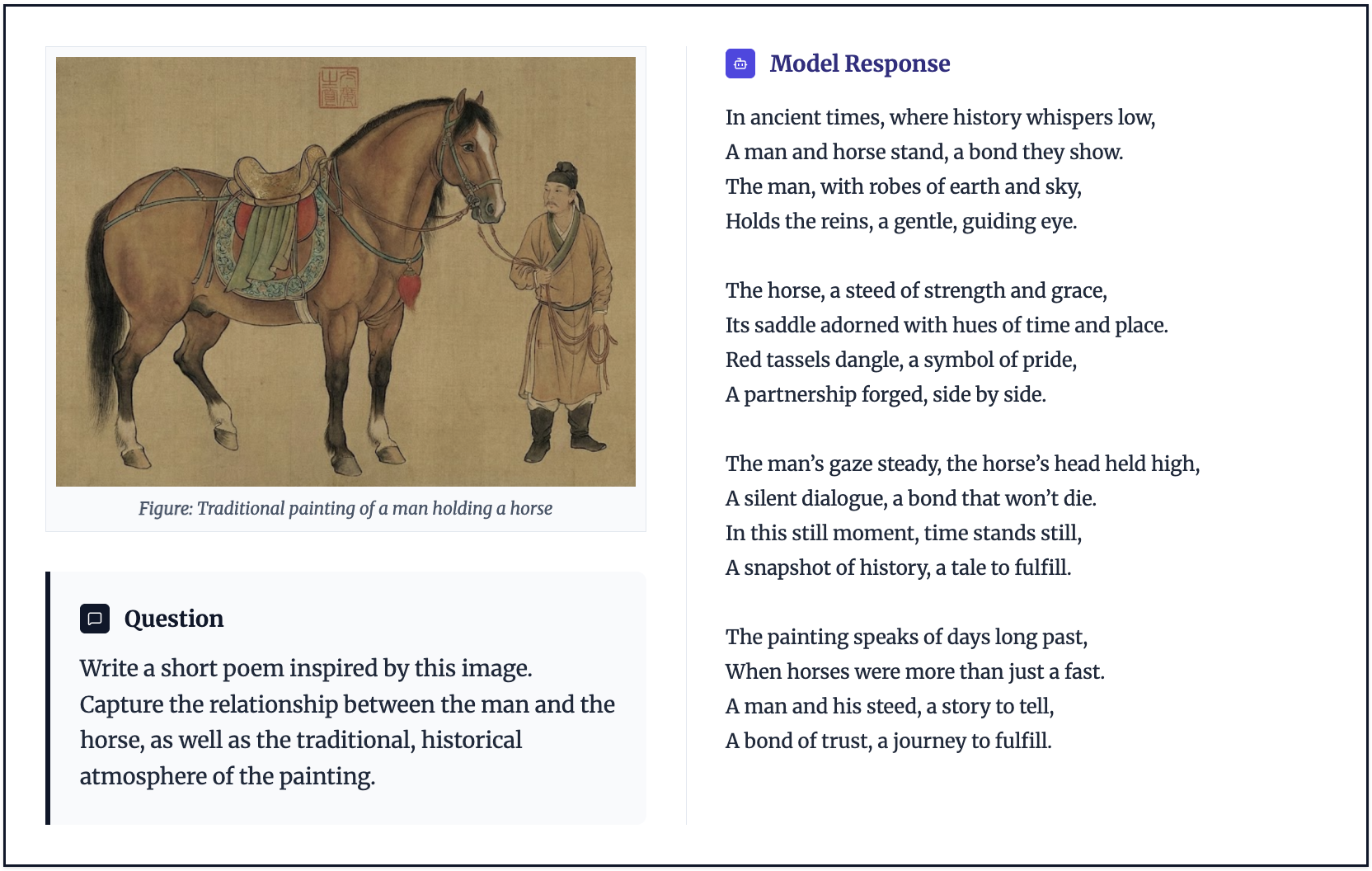} 
\caption{Case study of artistic interpretation and poetic creative writing. The input figure caption is for illustration only and is not provided to the model.
}
\label{fig:case_art}
\end{figure*}

\subsection{Image and Video Case Study}

\noindent\textbf{Code Writing.} 
Figure~\ref{fig:case_code} showcases Penguin's capacity for complex algorithmic reasoning and executable code generation. Given a visual input of a ``Hard''-level competitive programming problem, the model does not merely transcribe the text; it formulates a rigorous dynamic programming (DP) formulation. The output details a step-by-step state transition logic, systematically handles edge cases for wildcard pattern matching, and generates syntactically correct Python code. This demonstrates a strong alignment between visual problem comprehension and formal symbolic reasoning.

\noindent\textbf{OCR and Document Understanding.} 
Figure~\ref{fig:case_ocr} visualizes the model's robustness in dense document understanding and layout-aware optical character recognition (OCR). Notably, despite Penguin adopting a vision encoder initialized purely from a text-based LLM, it still demonstrates profound fine-grained image understanding capabilities. Even with severe visual degradation, archaic typography, and complex spatial arrangements inherent to historical documents, Penguin executes high-fidelity text extraction. It successfully preserves the structural integrity of the document, maintaining precise left-to-right and top-to-bottom reading orders, which is a critical prerequisite for accurate downstream semantic parsing.

\noindent\textbf{Chart Image Understanding.} 
As illustrated in Figure~\ref{fig:case_chart}, Penguin exhibits advanced quantitative visual reasoning capabilities. When tasked with interpreting multivariate line charts, the model reliably extracts fine-grained data points—such as identifying global minima within a specific temporal window. Furthermore, it demonstrates higher-order comparative analysis by evaluating the relative volatility and stability across distinct data distributions over a 50-year span, highlighting its proficiency in structured data extraction and multi-step trend synthesis from non-Euclidean visual representations.

\noindent\textbf{Image Creative Writing.} 
In the domain of cross-modal generation, Penguin extends beyond explicit object detection to capture implicit semantic nuances and artistic styles. As demonstrated in Figure~\ref{fig:case_art}, when presented with a traditional painting of a man holding a horse, the model effectively parses both the concrete visual elements and the overarching historical atmosphere. It seamlessly translates specific visual details—such as the red tassels adorning the horse's saddle and the steady gaze between the figures—into an evocative, multi-stanza poem. This highlights Penguin's zero-shot ability to synthesize visual cues, interpret the abstract bond between subjects, and transform visual aesthetics into coherent, contextually aligned creative writing.

\noindent\textbf{Video Understanding and Temporal Grounding.} 
Figure~\ref{fig:case_video} and~\ref{fig:case_video1} evaluate the model on long-context video comprehension and fine-grained temporal localization. Penguin effectively processes extended visual sequences to construct comprehensive global narratives, capturing both high-level thematic shifts and fine-grained scene details. Crucially, the model demonstrates precise temporal grounding capabilities, successfully mapping specific semantic queries—such as architectural landmarks or key events—to exact timestamp intervals within a continuous 300-second video stream, indicating highly robust spatiotemporal representation learning.

\begin{figure*}[tbp] 
\centering
\includegraphics[width=1.\linewidth]{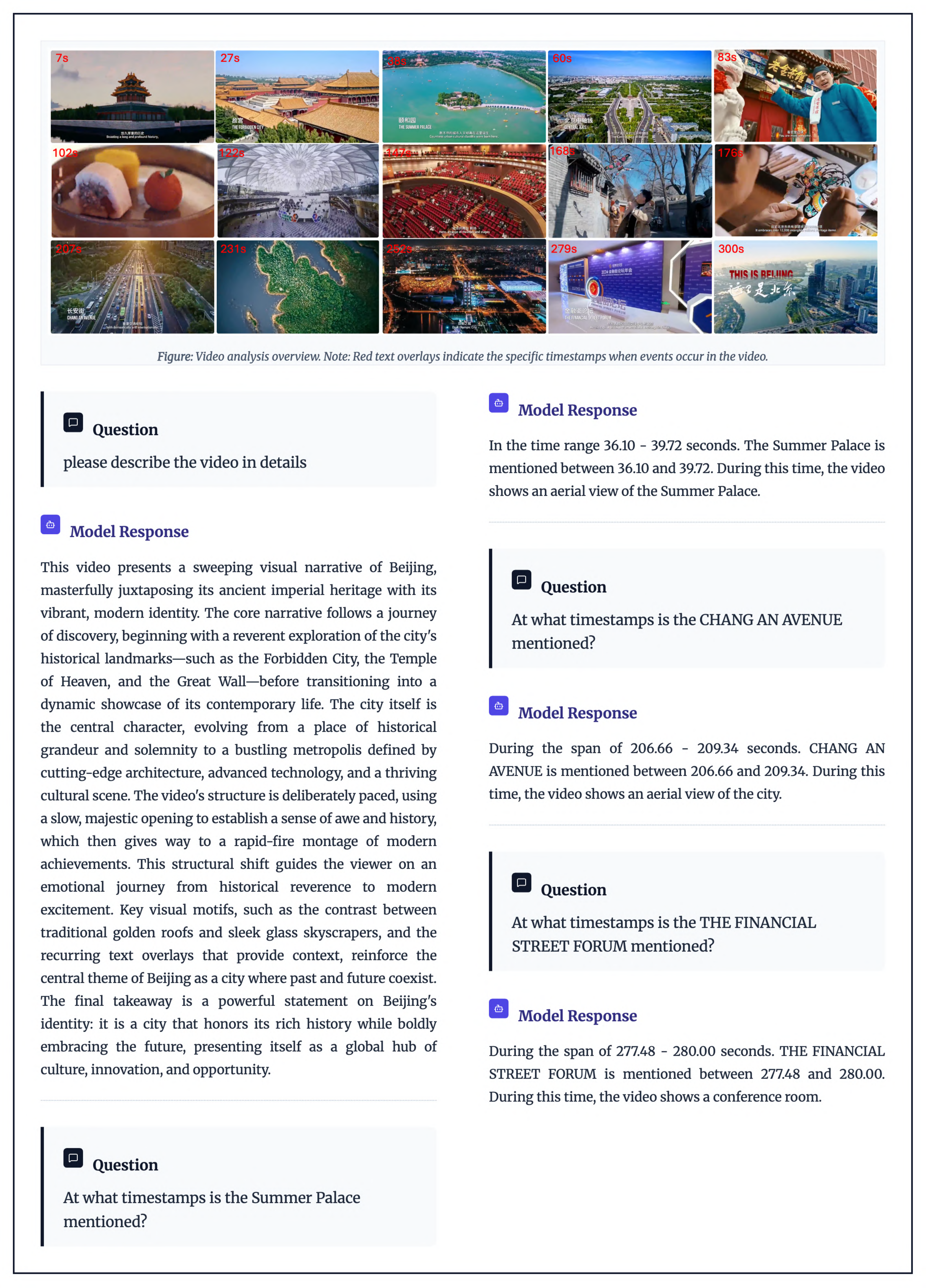} 
\caption{Case study of comprehensive video understanding and temporal grounding. The input figure caption is for illustration only.
}
\vspace{-0.1in}
\label{fig:case_video}
\end{figure*}

\begin{figure*}[tbp] 
\centering
\includegraphics[width=1.\linewidth]{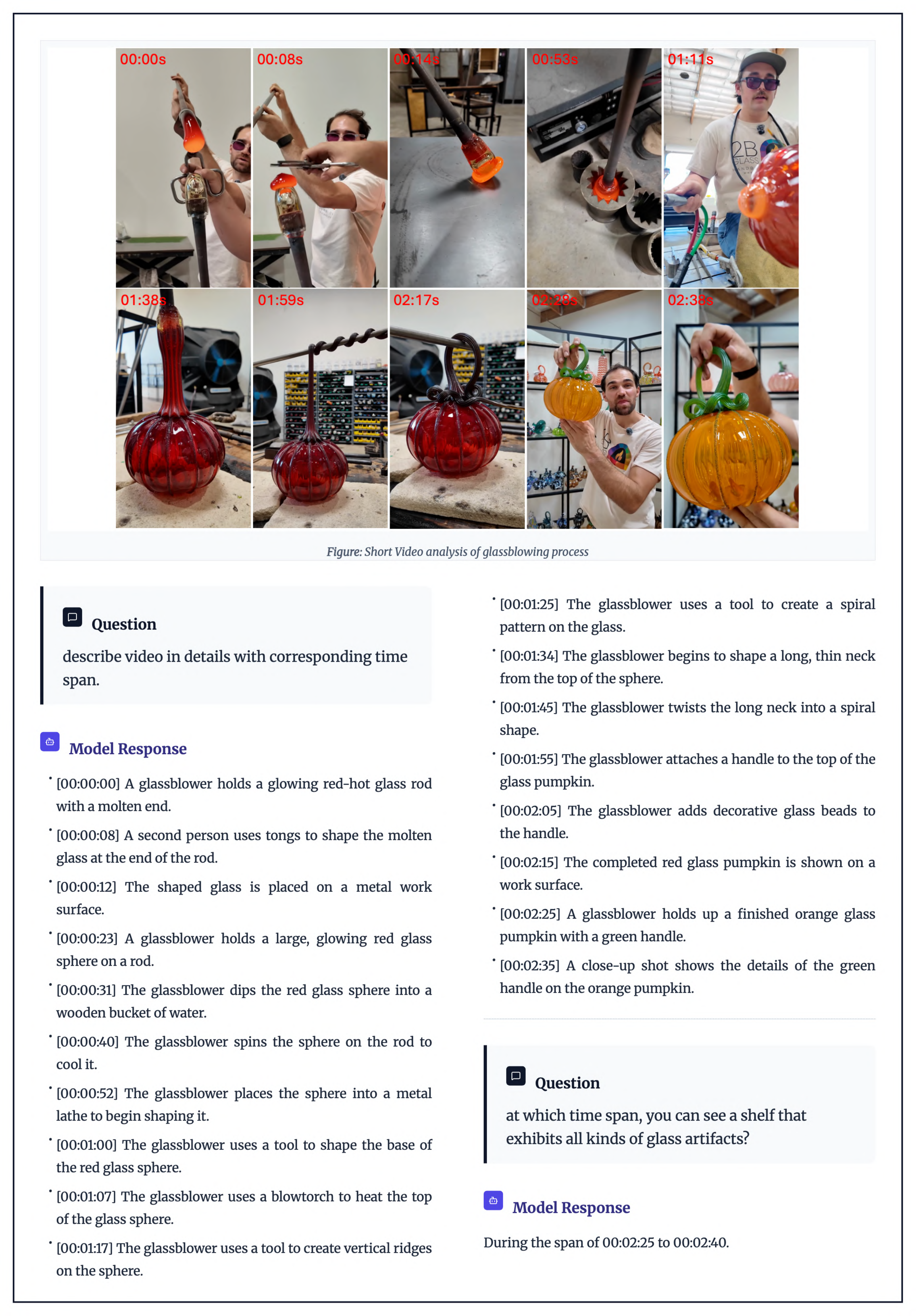} 
\vspace{-0.2in}
\caption{Case study of short video understanding and temporal grounding. The input figure caption is for illustration only and is not provided to the model.
}
\label{fig:case_video1}
\end{figure*}

\section{Related Work}

\paragraph{Vision-centric Multimodal-LLM} The landscape of VLMs has evolved rapidly over the past two years, marked by a definitive shift toward unified, high-resolution, and agentic multimodal systems. Leading proprietary models have pioneered end-to-end processing; notably, GPT-4o~\citep{hurst2024gpt} introduced autoregressive omni-modal capabilities across text, vision, and audio natively, while the Gemini series~\citep{comanici2025gemini} significantly expanded context windows to process hours of video alongside dynamic ``thinking'' budgets for complex reasoning. Concurrently, open-source models have made dramatic strides in architectural efficiency and capability scaling. The LLaVA-OneVision~\citep{li2024llavaonevision} and LLaVA-Video~\citep{zhang2024llava} frameworks transitioned away from task-specialized pipelines toward scenario-agnostic fusion for single-image, multi-image, and video comprehension. Gemma 3~\citep{team2025gemma} introduced efficient, long-context multimodal reasoning with reduced KV-cache memory overhead, democratizing access to massive context windows. To enhance temporal modeling and fine-grained visual grounding, the Molmo series~\citep{deitke2025molmo,clark2026molmo2} introduced robust spatial-temporal pointing and tracking mechanisms, while VideoLLaMA 3~\citep{zhang2025videollama} adopted a strictly vision-centric training paradigm leveraging high-quality image-text data for superior video understanding. Pushing the boundaries of open-source scaling, Qwen3-VL~\citep{Bai2025Qwen3VLTR} utilizes interleaved spatial-temporal representations natively supporting up to 256K tokens, and InternVL 3.5~\citep{wang2025internvl3} leverages Decoupled Vision-Language Deployment (DvD) and Cascade Reinforcement Learning to achieve enhanced reasoning. Furthermore, the paradigm has increasingly shifted toward agent-centric interactions. Models such as GLM-4.6V~\citep{zeng2025glm} natively integrate multimodal function calling and frontend visual editing, paralleling the Claude 4~\citep{claude4} series, which exhibit state-of-the-art autonomous computer use, long-horizon tool execution, and complex multi-system debugging.

\paragraph{Encoder Design}
Early vision–language models typically adopt vision encoders initialized from pretrained CLIP models~\citep{radford2021learning}, while some works instead pretrain their own ViT backbones on large-scale proprietary datasets using contrastive objectives~\citep{guo2025seed1}. Subsequent studies replace CLIP with SigLIP~\citep{Zhai2023SigmoidLF,tschannen2025siglip,zhang2025videollama} to improve language–semantic alignment during pretraining. Despite these differences, most approaches follow a similar paradigm: training a ViT encoder on image–text paired data with contrastive losses. To further enhance visual perception, especially for fine-grained understanding, several works incorporate SAM-based~\citep{kirillov2023segment} encoders to inject inductive biases from semantic segmentation, enabling a clearer separation between global semantic reasoning and localized visual tasks such as text reading or small-object recognition~\citep{wu2024deepseek,kirillov2023segment,ravi2024sam,wei2025deepseek}. Additionally, there were also works like EAGLE that introduced Mix-of-Encoder for more comprehensive vision understanding~\citep{shi2024eagle}. Most recently, DeepSeekOCR2~\citep{wei2026deepseek} proposes an LLM-based encoder that combines a causal LLM with a lightweight SAM encoder~\citep{kirillov2023segment} for document processing, utilizing learnable sequential tokens for adaptation. However, this model remains unverified on general vision tasks, and its permutation mechanism risks compromising the spatial relationships within the image input.

Beyond the choice of vision encoder backbone, the handling of pixels and visual features remains an active research topic, as images and videos often exhibit varying resolutions and naïve resizing can degrade visual information. Early approaches adopt tiling strategies that partition inputs into fixed-size grids~\citep{liu2023improvedllava,Chen2024HowFA}, augmented with spatial positional encodings to preserve layout information. More recent methods operate directly at the patch level and incorporate RoPE to support dynamic resolution handling~\citep{wang2024qwen2,bai2025qwen2,Bai2025Qwen3VLTR}. In parallel, the design of the projection layer bridging vision encoders and language models has evolved. Q-Former or resampler-based approaches unify visual tokens via attention-based compression~\citep{yu2025minicpm,li2023blip2bl}, whereas newer works favor lightweight MLPs combined with pooling mechanisms, trading increased token counts for finer-grained visual representations.

\paragraph{Video Understanding.}
Extending vision language models to video requires addressing the prohibitive token length caused by dense frame sampling and strong temporal redundancy. Early approaches, such as LLaVA-Video~\citep{lin2023video}, adopt a fixed frame budget and apply simple MLP or pooling operations to compress frame-level features, closely mirroring image-based token handling. Despite its simplicity, this strategy remains one of the most robust and is still widely used in recent video MLLMs.
To better align with continuous video streams, subsequent works introduce temporal convolution or other learnable temporal fusion modules to aggregate information across frames~\citep{bai2025qwen2,Bai2025Qwen3VLTR,cheng2024videollama}. More recent studies focus on selectively modeling frames based on their importance. For example, SlowFast-LLaVA~\citep{xu2024slowfast} processes the same video through dual branches with low-frame-rate/high-resolution and high-frame-rate/low-resolution streams to capture complementary temporal and spatial cues. Keye-VL~1.5~\citep{yang2025kwai} further generalizes this idea by dynamically identifying slow (key) frames and fast (motion) frames based on pixel-wise similarity, enabling adaptive allocation of resolution and computation according to temporal relevance.


\section{Conclusion and Future Work}

\subsection{Conclusion}

In this work, we introduced \modelname-VL, a compact, vision-centric multimodal foundation model that bridges the gap between image and video understanding. We challenged the standard practice of relying on massive contrastive pretraining for vision encoders, demonstrating that its discriminative nature actively suppresses the fine-grained visual cues required for advanced reasoning. To address this, we proposed the~\modelname-Encoder, initialized directly from a text-only LLM architecture. We supported this architecture with a robust, first-principles methodology, detailing a three-stage training pipeline (encoder training, VLM pretraining, and SFT) and introducing highly effective data curation strategies for open-source datasets. Extensive evaluations confirm that \modelname-VL sets a new standard for parameter-efficient VLMs at the 2B and 8B scales, showcasing exceptional capabilities from document OCR to long-context temporal reasoning. Ultimately, our research proves that aligning a vision encoder’s initialization with an LLM's generative objectives provides a vastly superior and more data-efficient path than scaling disconnected contrastive pretraining.

\subsection{Future Work}

Building on the foundations established by \modelname, we identify several promising directions for future research aimed at addressing current limitations and further expanding the model’s capabilities.

\noindent\textbf{Real-time Inference Optimization.} It is also crucial to address the real-time optimization of VLMs, enabling low-latency, interactive operation where perception, reasoning, and action must be performed under strict time and resource constraints. While current VLMs are typically optimized for offline accuracy, real-world applications (including embodied agents, GUI automation, and assistive systems) require adaptive computation that responds to dynamic inputs in real time. Future research may explore adaptive inference strategies, including early exiting, token/region-level sparsity, and dynamic resolution or frame-rate control, allowing the model to trade accuracy for latency on demand. In addition, incremental and streaming multimodal processing, where visual and linguistic representations are updated continuously rather than recomputed, can significantly reduce redundant computation. Finally, hardware-aware training, on-device distillation, and joint optimization across perception and reasoning modules will be critical to enabling VLMs that remain responsive, efficient, and robust in real-time interactive environments.

\noindent\textbf{Advanced Post-Training Techniques.} A key direction for future work is to move beyond SFT and explore reinforcement learning (RL)–based post-training for VLMs. While SFT provides strong initialization for general vision–language alignment, it is inherently limited by static annotations and cannot fully capture long-horizon objectives, interaction dynamics, or implicit user preferences. RL post-training offers a principled framework to optimize VLMs with respect to task-level rewards, enabling the model to refine decision-making, action grounding, and multimodal reasoning through trial-and-error interaction. Future research may investigate scalable reward modeling for vision–language tasks, environment-driven feedback from GUI or embodied settings, and hybrid SFT–RL pipelines that balance stability with exploration. Such RL-based post-training has the potential to significantly enhance robustness, adaptability, and agentic behavior, particularly in interactive and goal-oriented applications.

\noindent\textbf{Agentic Use.} Another important application is to extend our VLM beyond general-purpose vision understanding toward agentic visual-language systems, with a particular emphasis on GUI agents and computer-use scenarios. While the current model excels at perception and semantic reasoning over images, real-world deployment increasingly requires models that can interpret visual interfaces, decompose user intents, and execute multi-step interactions within dynamic graphical environments. Future research will focus on grounding visual understanding in actionable affordances, enabling the model to robustly perceive UI elements, track state changes, and plan long-horizon actions under partial observability. This direction also calls for tighter integration of perception, reasoning, and control, as well as training paradigms that leverage interaction traces, self-feedback, and environment-aware supervision. Ultimately, advancing VLMs toward agentic computer-use capabilities will unlock more practical and autonomous systems that bridge visual understanding with real-world task execution.

\textbf{Acknowledgment}

We thank Meng Yu, Yiwen Shao, Xuanru Zhou, Yuxin Wang, Jiahong Li and Tianzi Wang for their support and discussion at Tencent AI Lab.

\bibliographystyle{plain}
\bibliography{paper}

\end{document}